\documentclass[10pt]{article} 
\usepackage{booktabs}
\usepackage[preprint]{tmlr}

\usepackage{amsmath,amsfonts,bm}

\def\eqref#1{equation~\ref{#1}}

\def\1{\bm{1}}

\def\ry{{\textnormal{y}}}

\def\rvx{{\mathbf{x}}}

\def\vf{{\bm{f}}}

\def\vw{{\bm{w}}}
\def\vx{{\bm{x}}}

\def\vz{{\bm{z}}}

\DeclareMathAlphabet{\mathsfit}{\encodingdefault}{\sfdefault}{m}{sl}
\SetMathAlphabet{\mathsfit}{bold}{\encodingdefault}{\sfdefault}{bx}{n}

\DeclareMathOperator*{\argmax}{arg\,max}

\providecommand{\ry}{\textnormal{y}}      
\providecommand{\rsm}{\textnormal{m}}     
\providecommand{\vx}{\boldsymbol{x}}      
\providecommand{\rphi}{\phi}
\makeatletter
\@ifundefined{argmax}{\DeclareMathOperator*{\argmax}{arg\,max}}{}
\makeatother

\usepackage{hyperref}
\usepackage{url}
\usepackage{threeparttable}
\usepackage{graphicx}
\usepackage{subcaption}
\usepackage{amsmath}
\usepackage{xcolor}
\usepackage{multirow} 
\usepackage[normalem]{ulem}
\newcommand{\se}[1]{{\scriptsize$\pm$#1}}

\title{Learning to Defer under Expert Drift}

\author{\name Ziyao Mou\thanks{Currently affiliated with Tufts University. } \email ziyao.mou@tufts.edu \\
      \addr Department of Computer Science\\
      Johns Hopkins University
      \AND
      \name Andrea Wynn \email awynn13@jhu.edu \\
      \addr Department of Computer Science\\
      Johns Hopkins University
      \AND
      \name Eric Nalisnick \email nalisnick@jhu.edu\\
      \addr Department of Computer Science\\
      Johns Hopkins University}

\def\month{MM}  
\def\year{YYYY} 
\def\openreview{\url{https://openreview.net/forum?id=XXXX}} 

\begin{document}

\maketitle

\begin{abstract}
The \textit{learning to defer} (L2D) framework enables safety in machine learning and AI systems by allocating difficult or critical decisions to a human expert. However, most prior work on L2D assumes fixed expert reliability, overlooking temporal fluctuations in human performance due to factors such as fatigue and cognitive biases. Humans commonly exhibit such substantial, systematic performance shifts across a wide range of domains, from high-stakes settings such as radiology, aviation, and judicial decision-making, to everyday tasks such as crowdsourced image annotation. 
We propose \textit{Expert-Drift-Adapted Learning-to-Defer} (EDA-L2D), a novel expert-aware L2D framework that explicitly models these temporal shifts by conditioning the deferral head on recent histories of expert outcomes via a sequence model. This history-aware approach enables adaptive deferral policies that anticipate reliability fluctuations and better allocate decisions between human and machine over time. We provide comprehensive experiments across three diverse tasks—image classification, medical diagnosis, and hate speech detection—including generalization analyses under realistic deferral conditions and irregular shifts in expert reliability. Our results show that EDA-L2D adapts more effectively than prior L2D methods to diverse temporal patterns of expert reliability, resulting in improved system performance and more effective deferral decisions. Code is available at \url{https://github.com/ZiyaoMou/L2D-Expert-Drift}.
\end{abstract}

\section{Introduction}

\textit{Hybrid intelligent} (HI) systems combine human and machine intelligence to achieve goals unattainable by either alone, emphasizing complementarity over replacement \citep{Kamar2016HybridIntelligence,Dellermann2019HybridIntelligence,Akata2020ResearchAgendaHI}. One popular HI paradigm is \textit{learning to defer} (L2D): the system learns not only to predict a label but also to decide when to hand off the decision to a human expert. The original formulation of L2D shows that accounting for the expert's strengths and biases can improve overall accuracy and fairness relative to a static rejection framework \citep{madras2018predictresponsiblyimprovingfairness}. 

However, most existing L2D systems assume that expert behavior is fixed or stationary. In practice, human performance often varies significantly over time due to fatigue \citep{Figalova2024FatigueUnderloadL3, Peukert2023ATCFatigueDynamics, Pan2022PilotsFatiguefNIRS, bruni2012factors} and various cognitive biases \citep{Legler2025RepetitionBias, Urai2019ChoiceHistory, vandeWouw2024BiasedExpectations}. These changes can be large enough to matter operationally. For example, in a workshop-based mammography interpretation study involving 197 radiologists, specificity decreased from 82.74\% in the early morning to 71.70\% in the late morning, a decline of more than 11 percentage points \citep{alshabibi2020impact}. Even in crowdsourced image annotation, our analysis of CIFAR-10H \citep{peterson2019humanuncertaintymakesclassification} identifies 9 annotators who passed the dataset's official attention checks but nevertheless exhibited accuracy declines of more than 10 percentage points between the first and second halves of their annotation sessions (Appendix~\ref{app:cifar10h-annotators}). That expert drift can be observed even in such a simple task, together with the findings above, motivates L2D models that can adapt to expert drift over time.

\begin{figure*}[t]
    \centering
    \makebox[\textwidth][l]{
    \hspace*{-0.05\textwidth}

    \begin{subfigure}{0.48\textwidth}
        \centering
        \includegraphics[height=0.25\textheight,trim=0.5cm 0cm 0.5cm 0cm,clip]{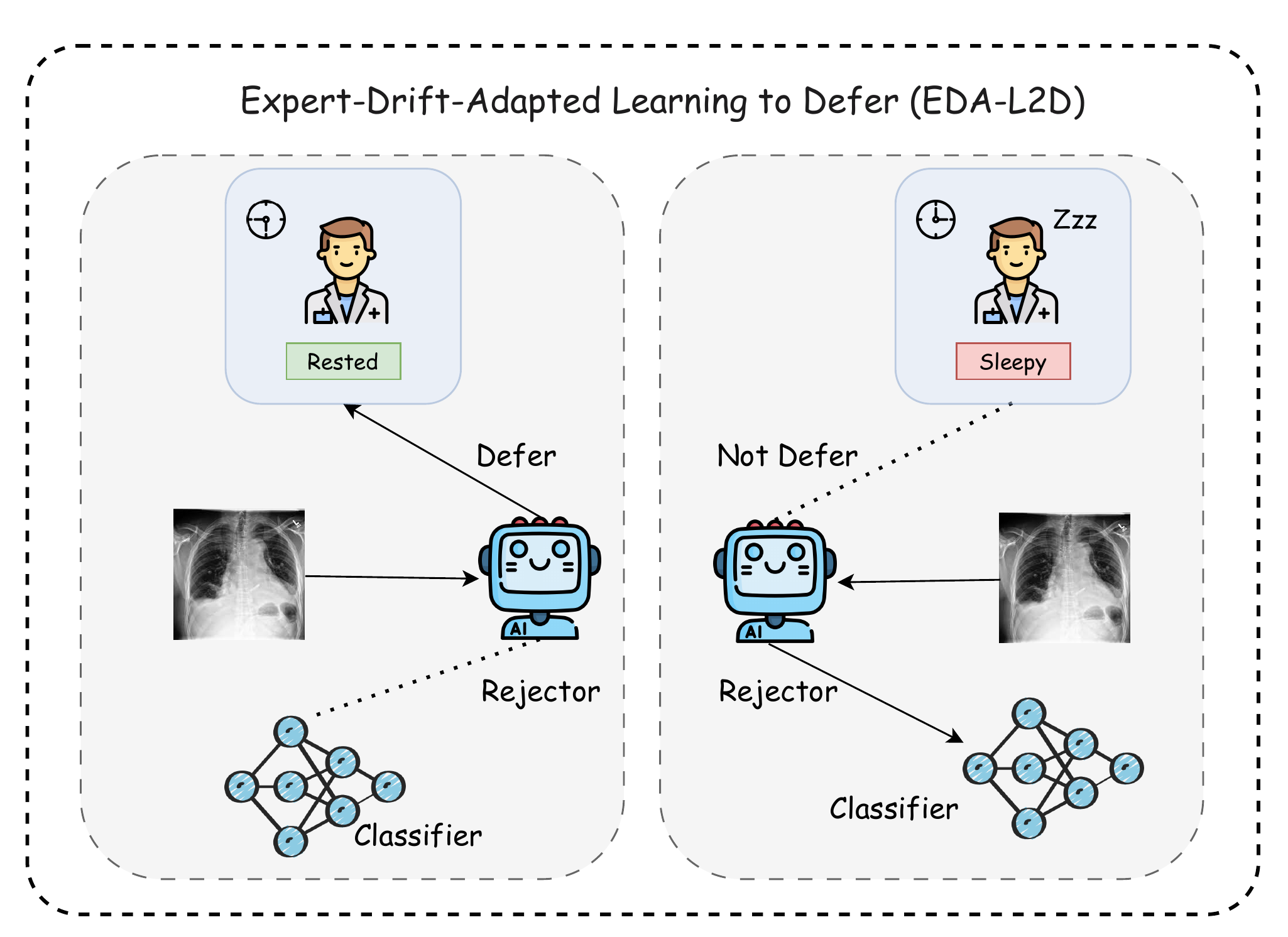}
        \caption{Learning to Defer with a Fatigued Expert}
        \label{fig:introduction}
    \end{subfigure}
    \begin{subfigure}{0.48\textwidth}
        \centering
        \includegraphics[height=0.25\textheight,trim=0.5cm 0cm 0.1cm 0cm,clip]{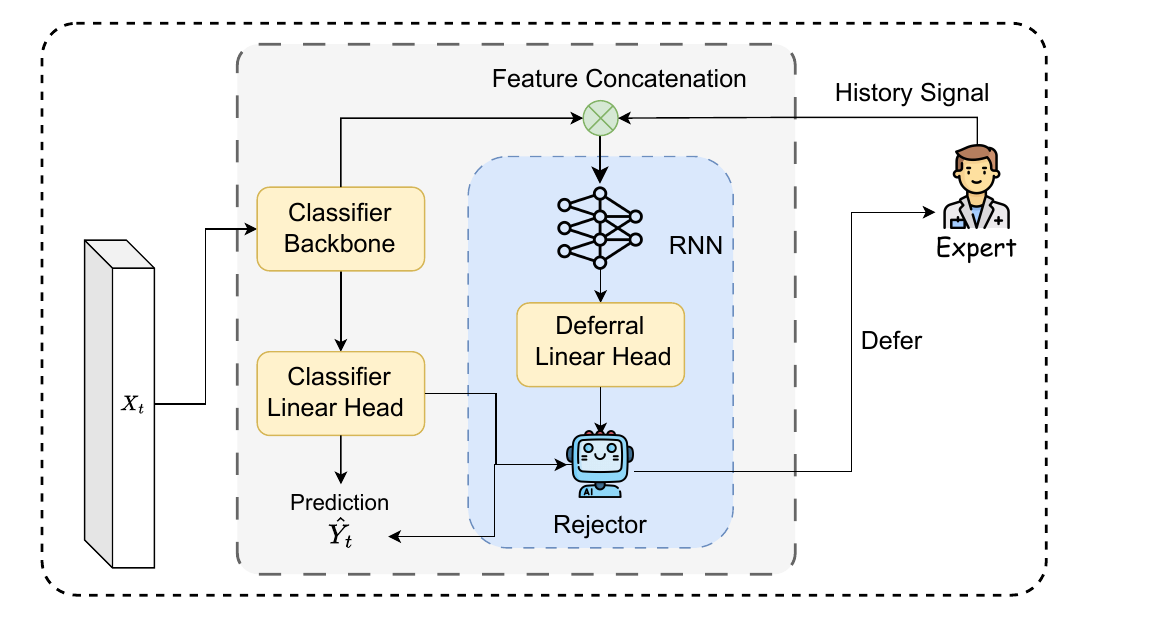}
        \caption{EDA-L2D System Architecture}
        \label{fig:model-intro}
    \end{subfigure}
    }
    
    \caption{(a) An example use case of our EDA-L2D framework in a setting with expert performance drift: a radiologist whose performance degrades over a workday. Our model monitors past outcomes to infer current expert reliability in real time, deferring to the expert when reliable and assigning cases to the model when the expert's performance declines. (b) Overview of our Expert-Drift-Adapted L2D framework. A streaming sequence of inputs is encoded by the backbone into features. At each timestep, the feature vector is concatenated with the expert’s most recently observed correctness, which is available only for inputs previously deferred to the expert, and passed to an RNN module. The classifier head outputs class logits and the defer head outputs a deferral score; a deferral gate then assigns the instance to either the expert or the classifier. Because the expert’s competence drifts over time, this design makes the L2D system time-aware and adaptive to temporal shifts.}
    \vspace{-0.5cm}
    \label{fig:genral}
\end{figure*}

We address this challenge by proposing a novel framework for L2D: \textit{Expert-Drift-Adapted Learning to Defer} (EDA-L2D).  Our framework explicitly models non-stationarity in expert performance.  We modify L2D's rejector sub-component to condition on short-horizon histories of expert decisions.  This allows the system to be aware of the expert's current behavior, and modify its deferral policy accordingly.  We illustrate our approach in Fig.~\ref{fig:introduction}. Concretely, we integrate a sequence model into the deferral head to explicitly model expert reliability over time.  We validate our approach across diverse tasks including medical image recognition, hate speech detection, and image classification. We demonstrate consistent gains in system accuracy over existing, time-agnostic L2D approaches.  In summary, our contributions are as follows:

\begin{itemize}
    \item We propose a novel \textit{Expert-Drift-Adapted Learning to Defer} (EDA-L2D) that explicitly models variance in human expert performance over time, allowing the system to flexibly adapt to realistic human experts. We propose our approach in \S\ref{sec:edal2d}, including a new time-aware learning objective (\S\ref{sec:learning-objective}) and model formulation (\S\ref{sec:expert-drift-aware-model}).
    \item We validate EDA-L2D on three real-world tasks that call for human--AI collaboration, each paired with a simulated expert whose accuracy drifts over time: image classification on \textsc{CIFAR-10} with drift curves taken from real \textsc{CIFAR-10H} annotators (\S\ref{sec:cifar10}), chest X-ray diagnosis on CheXpert with a radiologist's drift over a working week (\S\ref{sec:chexpert}), and hate speech detection with synthetic forms of drift (\S\ref{sec:hatespeech}). Both parameterizations of EDA-L2D consistently outperform standard L2D, and we further test their robustness to expert changes not seen in training and to delayed, noisy or missing feedback.

\end{itemize}

\section{Background}
We first describe the traditional L2D setting \citep{mozannar2021consistentestimatorslearningdefer}, in which the expert's abilities are assumed to be static and time-invariant.

\subsection{Data \& Models}
We will exclusively consider the problem of multi-class classification with a feature space \(\mathcal{X}\) and the label space \(\mathcal{Y}=\{1,\ldots,K\}\). Given some input \(x\in\mathcal{X}\), we define \(\mathbb{P}(\ry \mid x)\) as the unknown label-generating distribution and \(\mathbb{P}(\rsm \mid x)\)  as the unknown distribution over expert predictions \(m\in\mathcal{Y}\). Here, $\ry$ denotes the true label, whereas $\rsm$ denotes the prediction made by a fallible expert and may therefore differ from $\ry$. Their joint conditional distribution determines the expert's probability of being correct, $\mathbb{P}(\rsm=\ry\mid x)$, which is compared with the classifier's probability of being correct when deciding whether to defer. L2D is comprised of two sub-components: a \textit{classifier} \(h:\mathcal{X}\to\mathcal{Y}\) and a \textit{rejector} \(r:\mathcal{X}\to\{0,1\}\), which decides whether to make a prediction using the classifier (\(r(x)=0\)) or by deferring to the expert (\(r(x)=1\)).


\subsection{Classifier-Rejector Loss Function}
To train our L2D model, we need to fit both the rejector and classifier models. The classifier incurs a loss of zero (correct) or one (incorrect) when it makes a prediction. Similarly, the human expert incurs the same 0-1 loss when making a prediction (i.e.~$r(\vx)=1$). Combining these two 0-1 losses using the rejector model results in the combined classifier-rejector loss:
\begin{equation}\label{eq:0-1}\begin{split}
 L_{0-1}(h, r) =  \ \ \mathbb{E}_{\rvx, \ry, \rsm}\left[(1-r(\rvx)) \  \mathbb{I}[h(\rvx) \ne \ry] \ + \ r(\rvx) \ \mathbb{I}[\rsm \ne \ry] \right]
\end{split}
\end{equation} 
where $\mathbb{I}$ denotes an indicator function checking the prediction against the ground-truth label.  When minimizing this loss, the resulting Bayes optimal classifier and rejector satisfy: 
\begin{equation}\begin{split}\label{eq:Bayes_optimal_rej_clf_0-1_loss}
    h^{*}(\vx) = \argmax_{y \in \mathcal{Y}} \ \mathbb{P}(\ry = y | \vx), \quad
    r^{*}(\vx) = \mathbb{I}\left[\mathbb{P}(\rsm = \ry | \vx) \ge \max_{y \in \mathcal{Y}} \mathbb{P}(\ry = y | \vx) \right],
\end{split}
\end{equation} where $\mathbb{P}(\ry | \vx)$ represents the label probability under the data generating process, and $\mathbb{P}(\rsm = \ry | \vx)$ is the probability of the expert making the correct prediction.  The expert may have knowledge not available to the classifier, so they could outperform the Bayes optimal classifier.

\subsection{Softmax Surrogate Loss} 

A consistent surrogate loss for the above $L_{0-1}$ loss can be derived following the formulation from \cite{mozannar2021consistentestimatorslearningdefer}. First, we consider an augmented label space that includes both the label space $\mathcal{Y}$ and the rejection option $\bot$: $\mathcal{Y}^{\bot} := \mathcal{Y} \cup \{\bot\}$.  Secondly, for a class $k \in [1,K]$, let $g_{k}:\mathcal{X} \mapsto \mathbb{R}$, and let $g_{\bot}:\mathcal{X} \mapsto \mathbb{R}$ denote the rejection option. We can combine these $K+1$ with a loss resembling the cross-entropy loss for a softmax parameterization:
\begin{equation}\label{eq:sm_loss}\begin{split}
     \rphi_{\text{SM}}(g_{1}&,\ldots, g_{K}, g_{\bot}; \vx, y, m) =  -\log \left( \frac{\exp\{g_{y}(\vx)\}}{\sum_{y' \in \mathcal{Y}^{\bot}} \exp\{g_{y'}(\vx)\} }\right) - \mathbb{I}[m = y]  \ \log \left( \frac{\exp\{g_{\bot}(\vx)\}}{\sum_{y' \in \mathcal{Y}^{\bot}} \exp\{g_{y'}(\vx)\} }\right).
\end{split}
\end{equation} 
Here, the first term maximizes $g_{k}$ for the true label $k$.  The second term maximizes the rejection function $g_{\bot}$, but only when the expert's prediction is correct. At test time, the classifier takes the maximum over the classes: $\hat{y} = h(\vx) = \argmax_{k \in [1, K]} g_{k}(\vx)$.  Similarly, we formulate the rejection function as $r(\vx) = \mathbb{I}[ g_{\bot}(\vx) \ge \max_{k} g_{k}(\vx) ]$. The minimizers $g^{*}_{1},\ldots, g^{*}_{K}, g^{*}_{\bot}$ of $\rphi_{\text{SM}}$ also uniquely minimize the 0-1 loss from Equation \ref{eq:0-1}, $L_{0-1}(h, r) $. 
\section{Expert-Drift-Adapted Learning to Defer}
\label{sec:edal2d}
This section introduces our \textit{Expert-Drift-Adapted Learning to Defer} (EDA-L2D) framework, which extends L2D to account for non-stationarity in expert performance.  Accordingly, we derive the corresponding softmax surrogate loss for this setting.  We also propose multiple parameterizations, including one that uses a recurrent neural network to incorporate short-horizon temporal context into the deferral policy.

\subsection{Setting: Non-Stationary Expert}
\label{sec:expert-setting}
We consider the same setting as Section 2---L2D for multi-class classification---except now we assume the expert is non-stationary.  Thus, instead of a fixed distribution \(\mathbb{P}(\rsm \mid x)\), the expert's predictions are generated by \(\mathbb{P}_{t}(\rsm_{t} \mid x)\), where $t \in \mathcal{T} = \{1, \ldots, T\}$ is a discrete time index: the position of an input in the sequence of inputs the expert receives, such as the $t$-th image of an annotation session or the $t$-th case of a working day. Note that we are \emph{assuming the classification task itself is stationary}, i.e.~$\ry \sim \mathbb{P}(\ry \mid x)$, which still has no time dependence.

To ground the notation in an example, consider an L2D system for radiology.  The problem of forming a diagnosis or other prediction from the medical image itself is a static task, since we assume the distribution of patients and the mechanism that govern their health is not changing over time (or is at least changing so slowly as to be negligible, e.g.~drift in hospital patient demographics).  Yet the radiologist who is paired with the L2D system will change, at the very least becoming more tired and distracted over the course of a 10+ hour shift.  The expert drift may also be non-monotonic: perhaps her performance degrades over the course of a morning, but after lunch and a mid-day walk, the expert feels rejuvenated for a few more hours.

Returning to the technical details, our EDA-L2D system will have a classifier that is defined just as above: $h:\mathcal{X}\!\to\!\mathcal{Y}$.  Again, the classifier is time-invariant since the prediction task itself is not time dependent.  The difference, however, arises in the rejector:   $r:\mathcal{X} \times \mathcal{T} \mapsto \{0, 1\} $, meaning the rejector is a function of both the feature space $\mathcal{X}$ and time $\mathcal{T}$.  While in the simplest case $\mathcal{T}$ would be a scalar time index, we could also formulate $\mathcal{T}$ as a feature space that describes the current state of the expert (e.g.~tabular biomarkers).

\subsection{Learning Objective and Bayes Solutions}
\label{sec:learning-objective}
The learning objective is similar to that in Section 2 (\eqref{eq:0-1}), except now it takes the form of a summation over time:
\begin{equation}\label{eq:0-1-t}\begin{split}
 &L_{0-1}^{\mathcal{T}}(h, r)= 
 \sum_{t \in \mathcal{T}}\mathbb{E}_{\rvx, \ry, \rsm_t} \Big[(1-r(\rvx, t)) \, \mathbb{I}[h(\rvx) \ne \ry]  + \ r(\rvx, t) \, \mathbb{I}[\rsm_t \ne \ry] \Big].
\end{split}
\end{equation}  As the distribution of $\ry$ does not depend on time, the classifier's Bayes solution is just as before: $h^{*}(\vx) = \argmax_{y \in \mathcal{Y}} \ \mathbb{P}(\ry = y \mid \vx)$.  Yet the Bayes optimal rejector does change, becoming time-dependent like so: $$ r^{*}(\vx, t)  = \mathbb{I}\!\left[\mathbb{P}_{t}(\rsm_t = \ry \mid \vx) 
\;\ge\;  \max_{y \in \mathcal{Y}} \mathbb{P}(\ry = y \mid \vx) \right]. $$  The intuition is the same as above---compare the chance of expert correctness vs  classifier confidence for the modal label---yet now the expert correctness term is time dependent, $\mathbb{P}_{t}(\rsm_t = \ry \mid \vx)$.

\subsection{Implementations of the Softmax Surrogate Loss}  The derivation for the corresponding softmax surrogate loss follows fairly directly from \citet{mozannar2021consistentestimatorslearningdefer}'s original derivation.  Yet we consider two distinct parameterizations of the underlying model.  The first is a `per time step' version that treats the L2D problem independently at each time step.  We consider this the most naive extension of traditional L2D that yet still satisfies our motivating setting of expert drift.  We then go on to describe how to use a recurrent neural network (RNN) to parameterize the rejector, which allows the expert's previous predictions to help inform the deferral policy.  For all implementations, we assume we have access to a training set $\mathcal{D} = \left\{ \left\{ (\vx_{t,n}, y_{t,n}, m_{t,n}) \right\}_{n=1}^{N} \right\}_{t \in \mathcal{T}}$, which consists of $N$ feature-label-expert-prediction triplets at each of $|\mathcal{T}|$ time steps.  Again, we emphasize that only the underlying distribution of $m_{t,n}$ is time-varying, and the time indices on $\vx_{t,n}, y_{t,n}$ are there merely to `synchronize' the feature-label pairs with the expert predictions. 

\paragraph{Per-Time-Step Implementation and Loss}
The softmax surrogate in \eqref{eq:sm_loss} can be naturally extended to the drifting-expert setting by instantiating a traditional L2D model at each time step. Specifically, we define a model at each time step and train in locally for that time step via the surrogate:
\begin{equation}\label{eq:sm_loss_t}
\begin{aligned}
\rphi_{\text{SM}}^t(g_1, &..., g_K, g^{t}_\bot;\vx,y,m_t) = -\log \frac{\exp\{g_{y}(\vx)\}}{\sum_{y' \in \mathcal{Y}^{\bot}} \exp\{g_{y'}(\vx)\}} - \mathbb{I}[m_t = y] \log \frac{\exp\{g^{t}_{\bot}(\vx)\}}{\sum_{y' \in \mathcal{Y}^{\bot}} \exp\{g_{y'}(\vx)\}}.
\end{aligned}
\end{equation}  Notice that $g^{t}_\bot$ is specific to the current time step, but $g_1, \ldots, g_K$ could be shared across all time steps (since they encode the classifier).  As mentioned above, we consider this a naive implementation and will primarily use it to benchmark our second approach, described below.  Having $g^{t}_\bot(\rvx)$ only be a function of $\rvx$ means that it ignores patterns that could be detected by having information about the expert's behavior at previous time steps.
Moreover, the per-step parameterization has no dedicated defer head for time steps beyond the training horizon. Applying it to longer sequences therefore requires an extrapolation rule, such as reusing the final learned head, which we evaluate in Appendix~\ref{app:robustness}.

\paragraph{Loss for RNN-Based Model} We next consider a more sophisticated implementation that uses an RNN-based parameterization to incorporate information about the expert's previous predictions.  Here we use a surrogate loss defined across all time steps: 
\begin{equation}\label{eq:sm_loss_rnn}
\begin{aligned}
\rphi_{\text{SM}}^{\mathcal{T}}\left(g_1, ..., g_K, g_\bot;\{\vx_t,y_t,m_t\}_{t \in \mathcal{T}}\right) = & \sum_{t \in \mathcal{T}} -\log \frac{\exp\{g_{y_t}(\vx_t)\}}{\exp\{g_{\bot}(\vx_t , t)\} + \sum_{y' \in \mathcal{Y}} \exp\{g_{y'}(\vx_t )\}}
   \\ & - \mathbb{I}[m_t = y_t] \log \frac{\exp\{g_{\bot}(\vx_t , t)\}}{\exp\{g_{\bot}(\vx_t , t)\} + \sum_{y' \in \mathcal{Y}} \exp\{g_{y'}(\vx_t )\}}.
\end{aligned}
\end{equation} where here $g_{\bot}(\vx_t , t)$ is parameterized with an RNN so that it can leverage information from previous time steps.  While we have left the time feature $t$ vague, this will contain the expert's previous prediction $\rsm_{t-1}$ when it is available.  We next describe the neural architecture in more detail.

\paragraph{RNN Architecture}
\label{sec:expert-drift-aware-model}
The RNN first takes as input the feature vector $\rvx_t$, producing a hidden representation $\vf_t = \psi_{0}(\vx_t)$.  For instance, if $\rvx_t$ is an image, then $\psi_0(\cdot)$ could be a convolutional NN.  The feature representation is then passed into a recurrent module: $\vz_t = \psi_{1}\!\bigl(\vz_{t-1},\, \vf_t,\, \hat m_{t-1}\bigr)$. During training we use teacher forcing: $\hat m_{t-1} = \mathbb{I}[m_{t-1} = y_{t-1}]$ is the expert's true correctness at every step. At test time this is not available. The expert is consulted only on the inputs the system defers, so their correctness can only be observed for those inputs, for example when the expert's decisions are later verified. We therefore evaluate under \emph{realistic routing}: if input $t-1$ was deferred, $\hat m_{t-1}$ is the expert's observed correctness; otherwise, $\hat m_{t-1}$ keeps the last observed value, with a fixed initial value before the first deferral. Feedback in practice may also arrive late, be recorded incorrectly, or be missing, and we study these cases in Section~\ref{sec:cifar10} and Appendix~\ref{app:robust-imperfect-fb}. Lastly, the $g$-functions found in Equation \ref{eq:sm_loss_rnn} are produced by a final linear transformation: $g_k(\vx_t) = \vw_k^\top \vz_t \quad ( k\in [1, K])$ and $g_{\bot}(\rvx_t) = \vw_{\bot}^\top \vz_t$.  The predicted label is chosen via $\arg\max_k g_k(\rvx_t)$.  $g_{\bot}(\vx_t)$ quantifies the preference for deferral, with the system deferring if $g_{\bot}(\vx_t) \ge \max_k g_k(\rvx_t)$.

\section{Related Work}
\label{sec: related}
\paragraph{Learning to Defer (L2D)} Early work on classifiers that can choose to reject or abstain instead of making a prediction \citep{chow-decision-functions} inspired more recent work on how to model what happens after the classifier abstains \citep{madras2018predictresponsiblyimprovingfairness}, with \citet{mozannar2021consistentestimatorslearningdefer} deriving the first consistent surrogate loss for the problem. This work has been expanded in recent years to address several limitations, including mis-calibration \citep{verma2022calibratedlearningdeferonevsall, cao2023defensesoftmaxparametrizationcalibrated}, underfitting \citep{NEURIPS2022_bc8f76d9}, realizable consistency \citep{mozannar2023predictexactalgorithmslearning}, sample complexity \citep{charusaie2022sampleefficientlearningpredictors}, data scarcity \citep{hemmer2023learningdeferlimitedexpert}, and deferral to multiple experts \citep{verma2023learning}. However, these approaches do not consider the `messiness' of modeling \textit{human} experts---as we do---and could just as well be applied to a black-box API.

\paragraph{L2D for Sequences and Populations}  L2D has also been extended to the difficult settings of sequential tasks and encountering novel experts.  Regarding the former, \textit{Sequential Learning-to-Defer} \citep{Joshi2021SeqL2D} incorporates temporal dynamics by framing deferral as a sequential decision-making problem, employing model-based reinforcement learning.  This work, however, is primarily concerned with the sequential nature of the prediction task itself and not robustness to expert drift, which is our primary concern.  Regarding the latter, \textit{Learning to Defer to a Population} \citep{tailor2024learningdeferpopulationmetalearning} uses meta-learning to adapt to never-before-seen experts at test-time.  Their motivation is to be robust to novel experts (with the novelty not being a function of time), whereas we are concerned about modeling the temporal drift of a known expert.
 

\paragraph{Temporal Shifts in Human Performance} In many high-stakes domains, human performance shift over time (due to fatigue or other cognitive factors) are well-documented and widely studied. Fatigue is extremely common in high-stakes domains and long-horizon tasks, such as radiology \citep{bruni2012factors}, automated driving \citep{Figalova2024FatigueUnderloadL3}, air traffic control \citep{Peukert2023ATCFatigueDynamics}, and flying planes \citep{Pan2022PilotsFatiguefNIRS}. Beyond affecting surface-level performance on these tasks, fatigue physically affects the brain, causing reduced alertness and vigilance that can be measured directly, e.g. using EEG signals \citep{Figalova2024FatigueUnderloadL3}.  We extend L2D to be adaptive to these variations in human performance. 

\paragraph{L2D with Time-Varying Experts} Recent work has also begun to relax the assumption of a static expert. \citet{zhang2026fatigueawarelearningdeferconstrained} model expert accuracy as a parametric function of cumulative workload and learn a deferral policy with constrained reinforcement learning under a coverage budget. \citet{duy2026online} study online L2D with bandit feedback, where the set of available experts and their reliability may change across rounds. In contrast, EDA-L2D conditions the deferral decision on the expert's observed history of correctness, allowing it to track drift that follows no prescribed form.
\section{Experiments}

We evaluate our proposed EDA-L2D framework through a progression of controlled simulations. We train the two variants of EDA-L2D (per-step and RNN-based) on three benchmarks spanning distinct modalities and annotation regimes: \textbf{CIFAR-10} (image classification), \textbf{CheXpert} (chest X-ray classification) and \textbf{Hate Speech} (text moderation).  Our primary baseline is ``Naive L2D'', meaning that we apply a traditional L2D system that has no awareness of time, collapsing all training data and presenting it to the model without an information about time ordering.

\begin{table}[!b]
\centering
\caption{System accuracy (\%) on \textsc{CIFAR-10} with drifting \textsc{CIFAR-10H} annotators, averaged over the 50 time steps; mean $\pm$ standard error over 5 seeds, each evaluated on 20{,}000 Monte Carlo test sequences.}
\label{tab:cifar10}
\begin{threeparttable}
\small
\begin{tabular*}{\textwidth}{@{\extracolsep{\fill}}lccc}
\toprule
Method & Annotator 454 & Annotator 1004 & Annotator 54 \\
\midrule
Naive L2D & $90.38 \pm 0.05$ & $91.51 \pm 0.09$ & $93.93 \pm 0.09$ \\
EDA-L2D, Per-step param. & $\mathbf{93.83 \pm 0.08}$ & $93.52 \pm 0.07$ & $\mathbf{94.86 \pm 0.05}$ \\
EDA-L2D, RNN param. & $93.70 \pm 0.03$ & $\mathbf{93.66 \pm 0.07}$ & $\mathbf{94.84 \pm 0.06}$ \\
\quad delay 10\tnote{a} & $93.49 \pm 0.17$ & $93.64 \pm 0.07$ & $94.84 \pm 0.06$ \\
\quad no feedback\tnote{b} & $93.34 \pm 0.22$ & $93.62 \pm 0.07$ & $94.80 \pm 0.06$ \\
\bottomrule
\end{tabular*}
\begin{tablenotes}\footnotesize
\item[a] The observed outcomes reach the RNN only every 10 steps.
\item[b] The RNN receives no feedback about the expert's correctness.
\end{tablenotes}
\end{threeparttable}
\end{table}

\subsection{CIFAR-10 with Drifting CIFAR-10H Annotators}
\label{sec:cifar10}
\paragraph{Task and Data.}We study standard image classification on \textsc{CIFAR-10} \citep{Krizhevsky09learningmultiple}. The dataset contains 60{,}000 color images at $32\times32$ resolution drawn from 10 object categories (airplane, automobile, bird, cat, deer, dog, frog, horse, ship, truck), with 50{,}000 training images and 10{,}000 test images. Given an input image $x$, the system either predicts its class label $y\in\{1,\dots,10\}$ or defers to the expert ($\bot$). Images arrive in sequences, and the model can defer each one to a simulated expert whose accuracy drifts over time following accuracy curves of real annotators from \textsc{CIFAR-10H} \citep{peterson2019humanuncertaintymakesclassification}. We train on sequences of training images and evaluate on the test set in three ways: against simulated experts that follow each annotator's accuracy curve, against the annotators' real answers, and against changes to the expert or its feedback that are not seen during training.


\paragraph{Sequence Construction and Evaluation.} We cut the 50{,}000 \textsc{CIFAR-10} training images into consecutive sequences of length $T=50$, so that each image's time step is its position in its sequence. We then split the resulting 1{,}000 sequences at random into a training set containing 90\% of them and a validation set containing the remaining 10\%, so that each sequence belongs entirely to one set. The official 10{,}000-image test set is never seen during training and is used for two complementary evaluations. In the \emph{simulated evaluation}, which provides our main results, we generate 20{,}000 test sequences per trained model, drawing at every time step a test image uniformly at random and the expert's answer from the annotator's accuracy curve; this yields low-variance estimates of system accuracy at every step. In the \emph{real-answer evaluation}, we instead replay each annotator's 200 actual answers on the test images they labeled, in their original order. Each trial is placed at the time step it was mapped to when constructing the accuracy curve, which splits the 200 trials into four interleaved sequences of 50 steps (Appendix~\ref{app:cifar10-real-replay}). Unlike the simulated expert, whose correctness depends only on the time step, real answers also reflect which images the annotator found difficult, so this evaluation tests the simulator's central simplification as well as the transfer to real behaviour.

\paragraph{Models and Training.} All variants use a WideResNet-28-4 backbone with a $(K{+}1)$-way output layer, where the extra output is the deferral option, and are trained with the L2D surrogate loss. Naive L2D uses a single time-independent rejector. Per-step param.\ uses a single shared backbone and classifier head with 50 defer heads, one for each time step. The RNN-parameterized EDA-L2D passes the backbone features and the previous expert correctness through a recurrent network, using either GRU or LSTM cells, followed by a linear output layer, and trains the backbone and the recurrent network jointly. All models are trained for 200 epochs with SGD, Nesterov momentum $0.9$, weight decay $5\times10^{-4}$, a cosine learning-rate schedule, and a batch size of 200, which keeps the time steps of each sequence contiguous within a batch. We use 5 seeds for every reported result.

\begin{figure}[t]
\centering
\includegraphics[width=0.9\textwidth]{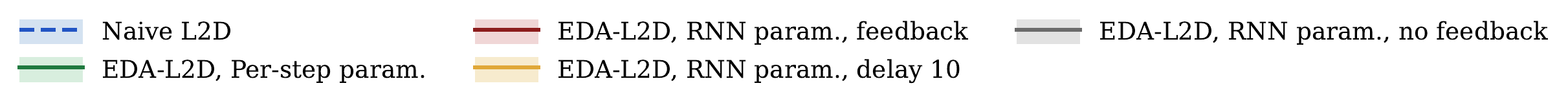}\\[2pt]
\begin{subfigure}{\textwidth}
\centering
\includegraphics[width=\textwidth]{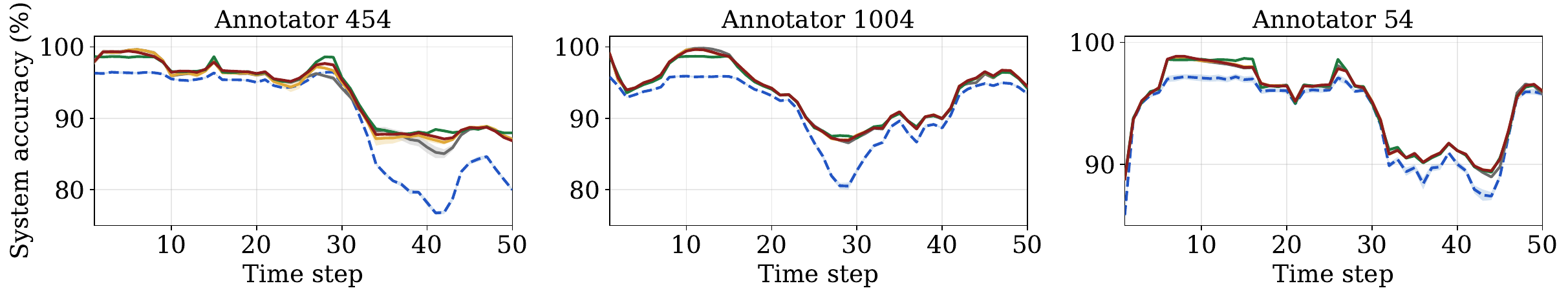}
\label{fig:cifar10-sysacc}
\end{subfigure}\\[-8pt]
\begin{subfigure}{\textwidth}
\centering
\includegraphics[width=\textwidth]{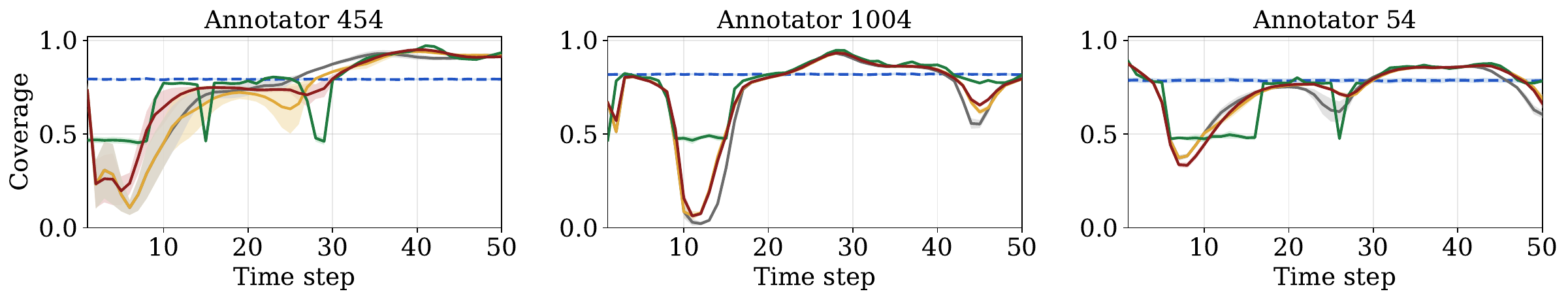}
\label{fig:cifar10-coverage}
\end{subfigure}\\[-12pt]
\caption{System accuracy and coverage (fraction of images not deferred) along the 50 time steps for the three \textsc{CIFAR-10H} annotators. Lines show the mean over 5 seeds and shaded bands $\pm$ one standard error. The RNN is evaluated under realistic routing, with the observed outcomes delivered immediately (feedback), every 10 steps (delay 10), or not at all (no feedback).}
\label{fig:cifar10-curves}
\end{figure}

\paragraph{Results.} Tab.~\ref{tab:cifar10} reports system accuracy averaged over the 50 time steps. Both EDA-L2D parameterizations clearly improve on Naive L2D for all three annotators (Per-step: $+3.45$, $+2.01$, $+0.94$ points; RNN: $+3.32$, $+2.15$, $+0.92$ for annotators 454, 1004 and 54), and both exceed the classifier and the expert alone. The two parameterizations perform similarly: Per-step is slightly better for annotator 454, the RNN for annotator 1004, and they tie for annotator 54, with differences of at most $0.14$ points. Fig.~\ref{fig:cifar10-curves} shows where the gains come from: Naive L2D applies the same deferral rule throughout, whereas both EDA-L2D variants lower their coverage when the expert is accurate and keep more images when the expert's accuracy drops, most visibly for annotator 454, whose accuracy falls sharply after step 30.

\paragraph{Real expert answers.} The gains carry over from simulated to real expert behaviour. Given Table \ref{tab:cifar10-real-answer}, when the annotators' actual answers are replayed, EDA-L2D, RNN param.\ improves on Naive L2D by $+3.8$, $+1.8$ and $+0.8$ points for annotators 454, 1004 and 54, close to the $+3.3$, $+2.2$ and $+0.9$ points obtained in the simulated evaluation. With only 200 trials per annotator, these estimates are noisier: the gains are significant for annotators 454 and 1004 (paired block bootstrap, 95\% intervals $[1.7, 6.0]$ and $[0.5, 3.3]$), whereas the smaller gain for annotator 54 cannot be resolved at this sample size.

\paragraph{Role of Historical Feedback.}
The RNN can draw on two sources of information about the expert: the temporal pattern of their accuracy, learned during training, and feedback on their past correctness, observed only for deferred images (Section~\ref{sec:expert-drift-aware-model}). When the expert follows the same curve at training and test time, the time step alone already reveals much of their accuracy, so feedback has little to add. Withholding it entirely costs at most $0.36$ points, and the RNN without feedback still gains $+2.96$, $+2.11$ and $+0.87$ points over Naive L2D (Tab.~\ref{tab:cifar10}). This contribution remains positive when feedback is delayed, noisy or only partly observed (Tab.~\ref{tab:cifar10-robust-feedback}). Feedback becomes more important when the expert departs from the pattern seen in training, because it is then the only signal of the change. During a temporary accuracy drop, its contribution grows with the depth of the drop, reaching $0.58$ points for annotator 454 at a drop of $0.15$ (Tab.~\ref{tab:cifar10-robust-dip-fb}). In short, the learned temporal pattern accounts for most of the gain when the expert behaves as expected, and feedback lets the model adapt when the expert does not (Appendix~\ref{app:robustness}).

\subsection{CheXpert with Radiologist Workweek Drift}
\label{sec:chexpert}

\begin{figure}[t]
\centering
\includegraphics[width=0.6\textwidth, trim=0 6 0 0, clip]{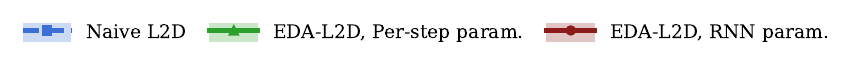}\\[-2pt]

\begin{minipage}[t]{0.495\textwidth}
  \centering
  {\scriptsize General cohort}\\[1pt]
  \includegraphics[width=\linewidth]{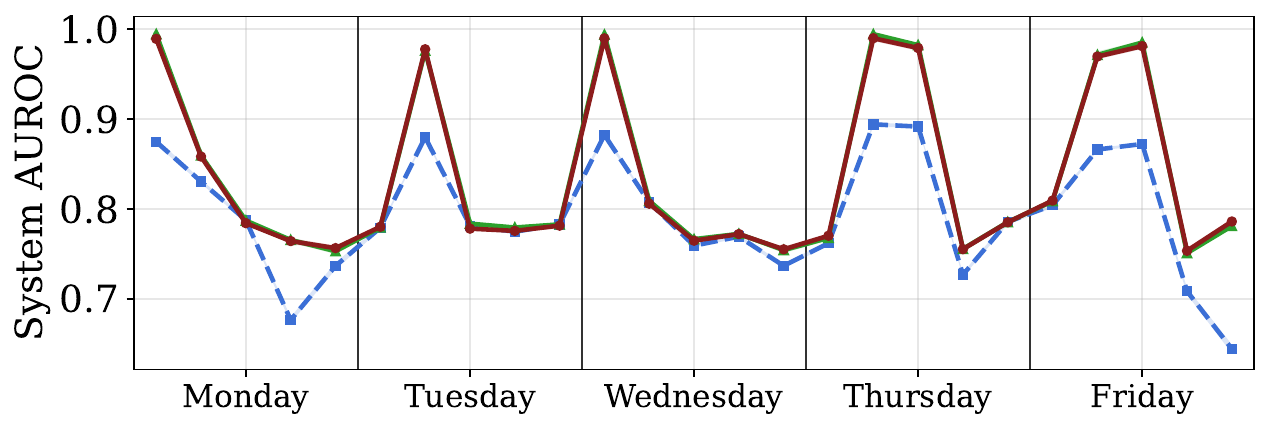}\\[2pt]
  \includegraphics[width=\linewidth]{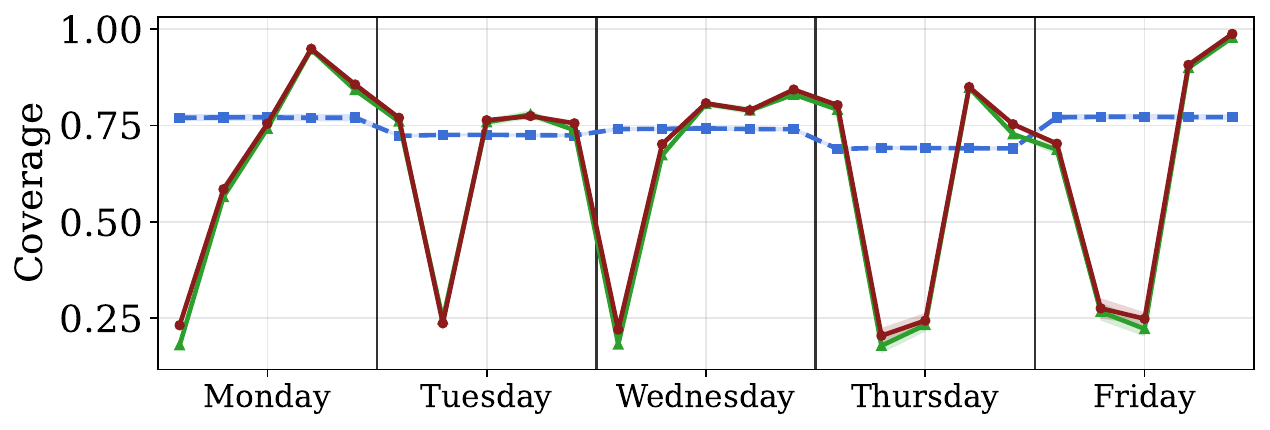}
\end{minipage}\hfill
\begin{minipage}[t]{0.495\textwidth}
  \centering
    {\scriptsize Emergency-department cohort}\\[1pt]
  \includegraphics[width=\linewidth]{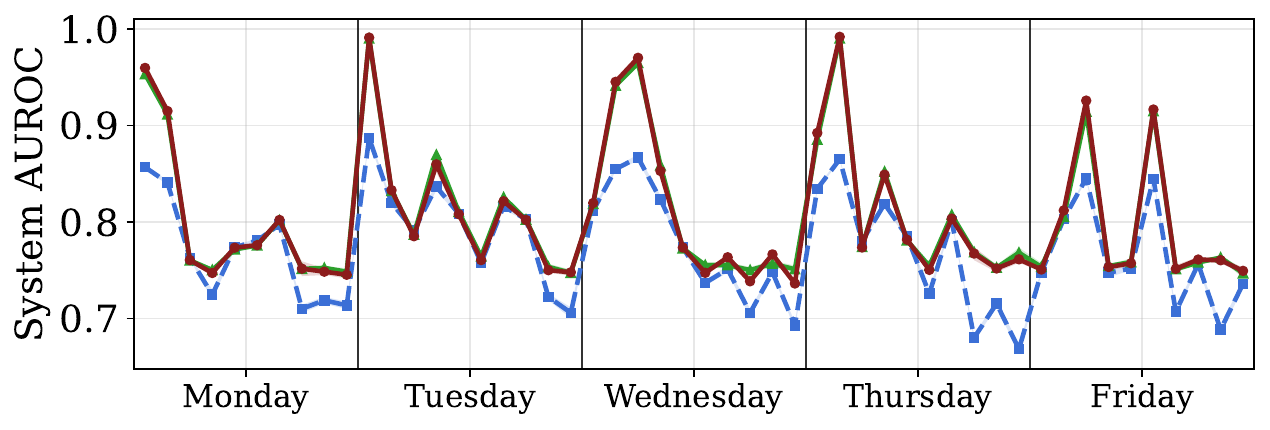}\\[2pt]
  \includegraphics[width=\linewidth]{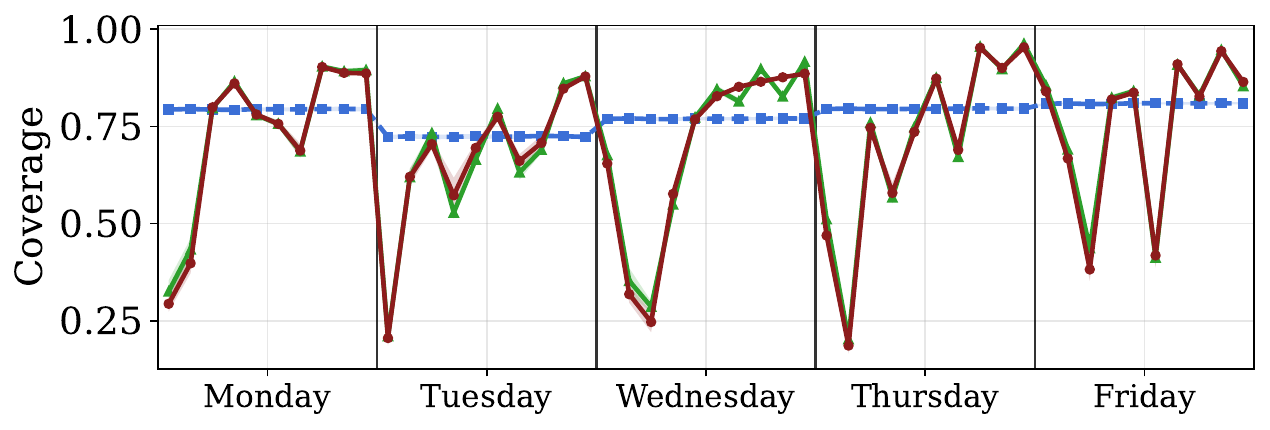}
\end{minipage}
\caption{CheXpert with one model per weekday, for the general cohort (left) and the emergency-department cohort (right). Lines show the mean over 5 seeds and shaded bands $\pm$ one standard error.}
\label{fig:chexpert-day}
\end{figure}
\paragraph{Task and Data.}
CheXpert~\citep{irvin2019chexpertlargechestradiograph} contains 224{,}316 chest radiographs from 65{,}240 patients at Stanford Hospital. We use the downsampled release and treat each observation as a one-vs-rest binary target. A binary mask excludes targets labeled uncertain. All images are converted to three channels, normalized, and resized to ImageNet-compatible resolution. For training augmentation, we apply random resized crops, horizontal flips, and random rotations of up to $15^\circ$. We split the data by patient into disjoint training, validation, and test sets containing 81\%, 9\%, and 10\% of patients, respectively.

\paragraph{Expert.}
We simulate an expert whose accuracy drifts over a working week. The temporal pattern comes from a radiology fatigue study~\citep{vosshenrich2021quantifying}: resident performance declines over each workday, partially recovers after midday breaks, and recovers less towards the end of the week. We digitize its mean trajectories for a general cohort (25 steps, 5 per day) and an emergency-department cohort (50 steps, 10 per day), rescale them to the range of balanced accuracies of individual CheXpert radiologists~\citep{irvin2019chexpertlargechestradiograph}, and amplify the drift so that it is pronounced enough to distinguish the methods. Appendix~\ref{app:chexpert-drift} gives the full construction and Figure~\ref{fig:chexpert-drift-curves} shows the resulting curves.

\paragraph{Setting.}
We consider two settings, and every method is trained under both. In the \emph{per-day} setting, each method is trained separately for each weekday on day-long sequences, which gives five models per method. Within a day, EDA-L2D, Per-step param.\ has one defer head per step, i.e., 5 heads in the general cohort and 10 in the emergency-department cohort. In the \emph{whole-week} setting, each method is a single model trained on week-long sequences that must cover the entire week, and EDA-L2D, Per-step param.\ has 25 or 50 defer heads, respectively. Because Naive L2D is also trained per day in the per-day setting, it already adapts to differences between weekdays, so the gains of EDA-L2D in this setting reflect its ability to track drift within a day. We present the per-day setting in the main text; results for the whole-week setting and the system AUROC along the week for individual findings are given in Appendix~\ref{app:chexpert-additional-results}.
\paragraph{Models and Training.}
All variants use a DenseNet-121 initialized with ImageNet weights, with one class output and one defer output per finding, and are fine-tuned with the same L2D loss from a classifier first trained without deferral. They differ only in the defer outputs: Naive L2D has a single one per finding, EDA-L2D, Per-step param.\ one per step, and EDA-L2D, RNN param.\ obtains them from a GRU over the image features and the expert's past correctness on each finding. Training details are given in Appendix~\ref{app:chexpert-hparams}.

\paragraph{Evaluation.}
For each finding, the system's score on an image is the expert's binary answer if the finding is deferred, and the model's predicted probability of the finding otherwise. We compute the AUROC of these mixed scores separately for each finding and time step, using only labeled cases, and report system AUROC as the average over findings and steps.

\paragraph{Results.}
Both parameterizations of EDA-L2D clearly outperform Naive L2D in both cohorts (Tables~\ref{tab:chexpert-day-general} and~\ref{tab:chexpert-day-emergency}), raising the average system AUROC from $0.792$ to $0.837$ and $0.836$ in the general cohort and from $0.774$ to $0.807$ and $0.806$ in the emergency-department cohort. The two parameterizations perform alike, and the gains are largest on findings for which the classifier alone is weak, such as Enlarged Cardiomediastinum and Fracture; Figure~\ref{fig:chexpert-findings-day} in Appendix~\ref{app:chexpert-additional-results} shows the system AUROC of these findings along the week. 
Figure~\ref{fig:chexpert-day} shows that EDA-L2D adapts its deferral to the expert's performance: its coverage falls when the expert is accurate and rises as the expert tires, whereas Naive L2D defers at an almost constant rate. As a result, EDA-L2D stays at or above Naive L2D throughout the week.

\begin{table}[htbp]
\centering
\caption{System AUROC on CheXpert, general cohort, one model per weekday, for three findings and averaged over all 14 findings; each value is averaged over the steps of the week, with mean $\pm$ standard error over 5 seeds. Bold: the best L2D method and any L2D method within one standard error of it.}
\label{tab:chexpert-day-general}
\small
\begin{tabular*}{\textwidth}{@{\extracolsep{\fill}}lcccc}
\toprule
Method & Enl.\ Cardiomediastinum & Consolidation & Fracture & Average (14 findings) \\
\midrule
Naive L2D & $0.680 \pm 0.004$ & $0.764 \pm 0.005$ & $0.736 \pm 0.008$ & $0.792 \pm 0.002$ \\
EDA-L2D, Per-step param. & $\mathbf{0.767 \pm 0.004}$ & $\mathbf{0.820 \pm 0.005}$ & $\mathbf{0.809 \pm 0.009}$ & $\mathbf{0.837 \pm 0.002}$ \\
EDA-L2D, RNN param. & $0.759 \pm 0.005$ & $\mathbf{0.819 \pm 0.001}$ & $\mathbf{0.806 \pm 0.006}$ & $\mathbf{0.836 \pm 0.001}$ \\
\bottomrule
\end{tabular*}
\end{table}

\begin{table}[!b]
\centering
\caption{System AUROC on CheXpert, emergency-department cohort, one model per weekday; same layout as Table~\ref{tab:chexpert-day-general}.}
\label{tab:chexpert-day-emergency}
\small
\begin{tabular*}{\textwidth}{@{\extracolsep{\fill}}lcccc}
\toprule
Method & Enl.\ Cardiomediastinum & Consolidation & Fracture & Average (14 findings) \\
\midrule
Naive L2D & $0.678 \pm 0.004$ & $0.744 \pm 0.004$ & $0.729 \pm 0.007$ & $0.774 \pm 0.001$ \\
EDA-L2D, Per-step param. & $\mathbf{0.727 \pm 0.007}$ & $\mathbf{0.797 \pm 0.003}$ & $\mathbf{0.769 \pm 0.012}$ & $\mathbf{0.807 \pm 0.001}$ \\
EDA-L2D, RNN param. & $\mathbf{0.729 \pm 0.005}$ & $0.790 \pm 0.006$ & $\mathbf{0.769 \pm 0.009}$ & $\mathbf{0.806 \pm 0.001}$ \\
\bottomrule
\end{tabular*}
\end{table}

\begin{figure}[t]
\centering
\includegraphics[width=0.7\textwidth]{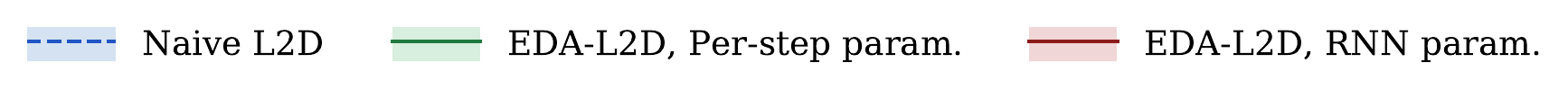}\\[-2pt]
\includegraphics[width=0.75\textwidth]{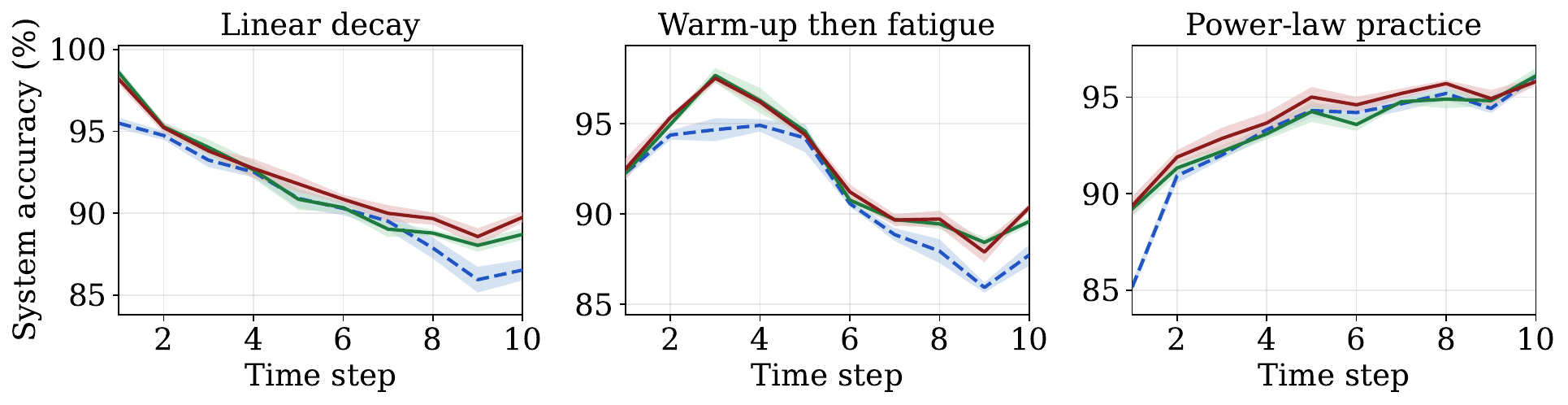}\\
\includegraphics[width=0.75\textwidth]{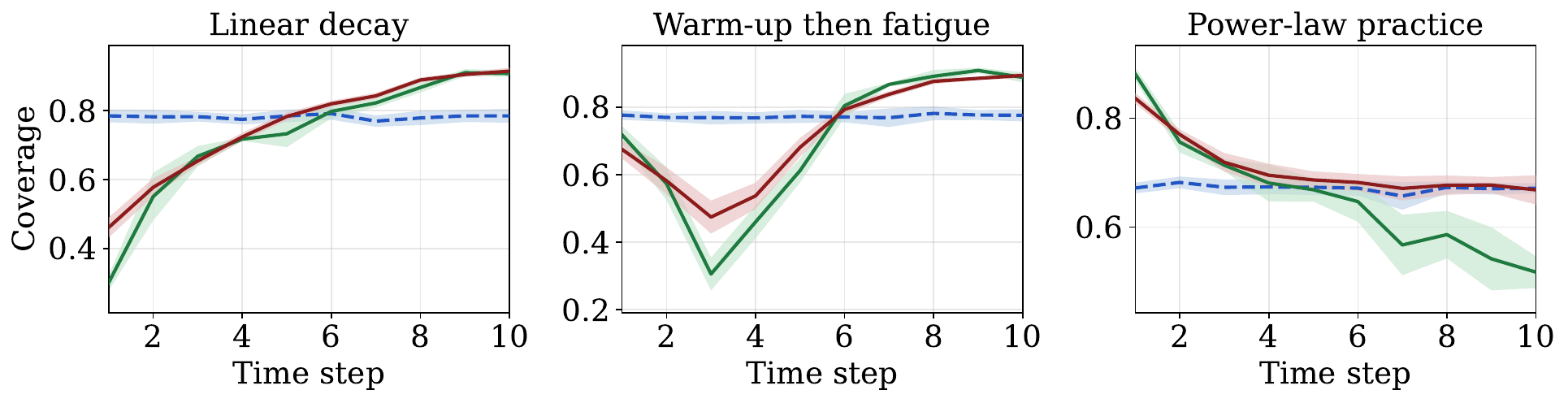}
\caption{System accuracy (top) and coverage (bottom) along the 10 time steps on hate speech, for the three synthetic experts. Lines show the mean over 5 seeds and shaded bands $\pm$ one standard error.}
\label{fig:hs-curves}
\end{figure}

\subsection{Hate Speech with Synthetic Expert Drift}
\label{sec:hatespeech}
\begin{table}[t]
  \centering
  \small
  \caption{Hate speech with synthetic expert drift: system accuracy (\%), averaged over the $T=10$ time steps (mean $\pm$ standard error over 5 seeds).}
  \label{tab:hs-synthetic}
  \begin{tabular*}{\textwidth}{@{\extracolsep{\fill}}lccc}
    \toprule
    Method & Linear decline & Warm-up then fatigue & Power-law practice \\
    \midrule
    Naive L2D                  & $90.70 \pm 0.17$ & $91.14 \pm 0.18$ & $93.03 \pm 0.06$ \\
    EDA-L2D, Per-step param.   & $91.62 \pm 0.21$ & $92.37 \pm 0.24$ & $93.43 \pm 0.15$ \\
    EDA-L2D, RNN param.        & $\mathbf{92.05 \pm 0.11}$ & $\mathbf{92.48 \pm 0.19}$ & $\mathbf{93.90 \pm 0.24}$ \\
    \bottomrule
  \end{tabular*}
\end{table}

\paragraph{Task and Data.} We study detection of abusive content on Twitter posts using the \citet{davidson2017automatedhatespeechdetection} corpus of 24{,}783 English tweets annotated into three mutually exclusive classes: \emph{hate speech}, \emph{offensive but not hate}, and \emph{neither}. To obtain a stream on which expert quality can change over time, we group the tweets, in the order of the released file, into consecutive blocks of $T=10$ tweets and treat each block as one sequence.

\paragraph{Synthetic experts.} The corpus records no annotator's accuracy over a controlled sequence, so we simulate the expert. At step $t \in \{1, \dots, T\}$, the expert answers a tweet correctly with probability $p_t$, independently of the tweet, and a miss earns no credit. With $u = (t-1)/(T-1) \in [0,1]$, we study three curves, each motivated by a mechanism reported for human annotators. The \emph{linear decline}, $p_t = p_{\mathrm{start}} - (p_{\mathrm{start}} - p_{\mathrm{end}})\,u$, is the simplest form of the drift observed in abusive-language annotation, where judgments shift between labelling sessions and depend on the conditions under which they are collected~\citep{abercrombie-etal-2023-temporal,kern-etal-2023-annotation}. The \emph{warm-up then fatigue} curve,
\[
p_t =
\begin{cases}
w_0 + (w_{\mathrm{peak}} - w_0)\,(u/\hat{u})^2, & u \le \hat{u},\\[2pt]
w_{\mathrm{base}} + \dfrac{w_{\mathrm{peak}} - w_{\mathrm{base}}}{1 + \exp\{K\,(u - \bar{u})\}}, & u > \hat{u},
\end{cases}
\]
follows the two-phase performance model of \citet{zhang2026fatigueawarelearningdeferconstrained}, in which performance first improves as a person settles into the task and then declines as fatigue accumulates. The \emph{power-law practice} curve, $p_t = p_\infty - (p_\infty - p_{\mathrm{start}})\,t^{-c}$, follows the power law of practice~\citep{Newell81} and drifts in the opposite direction to the linear decline. The curves are stylised test cases rather than fits to any annotator; each crosses the accuracy of the classifier alone at least once, so the better choice between classifier and expert changes over the sequence. Appendix~\ref{app:hs-curves} gives their parameters.

\paragraph{Backbone and training.} All methods use the same text classifier: a convolutional encoder over $100$-dimensional GloVe embeddings, with $300$ filters of widths $3$, $4$ and $5$, max pooling and dropout $0.5$. EDA-L2D, RNN param.\ adds a recurrent layer on top of the encoder features for the deferral decision. All methods are trained with the L2D loss using Adam at a constant learning rate of $10^{-3}$, without a learning-rate schedule, and batches of $64$ tweets ($64$ sequences for EDA-L2D, RNN param.). For each method we keep the epoch with the highest validation system accuracy and report the mean over five seeds. Further details are given in Appendix~\ref{app:imple-hatespeech}.

\paragraph{Results.} Table~\ref{tab:hs-synthetic} reports the system accuracy averaged over the ten time steps, and Figure~\ref{fig:hs-curves} shows system accuracy and coverage along the steps. Both parameterizations of EDA-L2D outperform Naive L2D on all three curves. EDA-L2D, RNN param.\ achieves the highest system accuracy, improving on Naive L2D by $1.35$ points on the linear decline, $1.33$ points on the warm-up-then-fatigue curve and $0.88$ points on the power-law practice curve. EDA-L2D, Per-step param.\ recovers most of this gain ($+0.92$, $+1.22$ and $+0.40$ points).

\section{Conclusion and Discussion}

We propose Expert-Drift-Adapted Learning to Defer (EDA-L2D), a framework that models temporal drift in human expert performance, a common phenomenon caused by factors such as fatigue and cognitive biases. Most L2D methods treat the expert as a fixed oracle, yet human reliability routinely changes over the course of work. EDA-L2D closes this gap by learning from the expert’s observed outcomes, so that its deferral decisions track the expert’s current reliability without assuming a particular form of drift. It incorporates time and the expert's history of correctness into the L2D model, learning time-varying representations of the expert and adapting its deferral decisions accordingly. Across image classification, chest X-ray diagnosis and hate speech detection, EDA-L2D consistently improves human-AI team performance over standard L2D. These gains largely hold under realistic conditions, where the expert's correctness is observed only for deferred cases and may be delayed, noisy or partial, and they persist to a considerable extent when the expert departs from the behavior seen in training.


 Our work complements several lines of L2D research discussed in Section~\ref{sec: related}, including sequential deferral \citep{Joshi2021SeqL2D}, deferral to unseen or changing experts \citep{tailor2024learningdeferpopulationmetalearning,duy2026online}, and fatigue-aware deferral \citep{zhang2026fatigueawarelearningdeferconstrained}. Building on these connections, our framework extends naturally to the multi-expert setting of \citet{verma2023learning}, with a separate temporal state per expert so that each case can be routed to the expert who is currently most reliable, and to populations of experts, where meta-learning drift patterns, as \citet{tailor2024learningdeferpopulationmetalearning} do for expert characteristics, would let a new expert's drift be inferred from few observations. Since the expert's correctness is observed only on deferred cases, a system that rarely defers cannot notice when the expert recovers; deferring occasionally to gather feedback, as the exploration in the bandit setting of \citet{duy2026online}, could address this.

Richer expert models would also help with effects that our current model does not capture. In our experiments, the expert's accuracy depends only on the time step, whereas fatigue effects may depend on case difficulty~\citep{krupinski2010long}. Real experts are also subject to content-dependent biases, such as anchoring~\citep{TURVEY2012495} and the gambler's fallacy in sequential decisions~\citep{Kovic04032019}, which a recurrent model of past correctness alone may not fully capture. Longitudinal data from real working sessions would allow these effects to be modeled and evaluated directly.

More broadly, our results suggest that human-AI collaboration should account for the dynamic nature of human performance: experts are not fixed oracles, and their reliability changes over the course of their work. We hope this work encourages learning-to-defer methods that adapt to these changes, so that human-AI teams remain effective as the people in them change.

\subsubsection*{Broader Impact Statement}
EDA-L2D aims to make human-AI teams more reliable by accounting for changes in expert performance, such as fatigue, which could reduce errors in settings like clinical reading and content moderation. It also carries risks. Records of an expert's correctness over time could be repurposed to monitor individual workers, so their use should be restricted to the deferral decision. When the model judges the expert to be less reliable, it takes over more cases, which can reduce human oversight if the model itself errs; experts should remain able to override it. Our experiments use simulated experts and have not been validated with real clinicians or moderators, which is necessary before any high-stakes deployment. Finally, the hate speech dataset contains known annotation biases that a deferral system may inherit.



\bibliography{main}

@article{Joshi2021SeqL2D,
  title = {Learning-to-defer for sequential medical decision-making under uncertainty},
  author = {Shalmali Joshi and Sonali Parbhoo and Finale Doshi-Velez},
  journal = {Transactions on Machine Learning Research},
  year = {2023},
  url = {https://arxiv.org/abs/2109.06312}
}

@inproceedings{cao2023defensesoftmaxparametrizationcalibrated,
  title = {In Defense of Softmax Parametrization for Calibrated and Consistent Learning to Defer},
  booktitle = {Advances in Neural Information Processing Systems},
  author = {Cao, Yuzhou and Mozannar, Hussein and Feng, Lei and Wei, Hongxin and An, Bo},
  editor = {Oh, A. and Naumann, T. and Globerson, A. and Saenko, K. and Hardt, M. and Levine, S.},
  year = 2023,
  volume = {36},
  pages = {38485--38503},
  publisher = {Curran Associates, Inc.},
  doi = {10.52202/075280-1671}
}

@InProceedings{verma2022calibratedlearningdeferonevsall,
  title = 	 {Calibrated Learning to Defer with One-vs-All Classifiers},
  author =       {Verma, Rajeev and Nalisnick, Eric},
  booktitle = 	 {Proceedings of the 39th International Conference on Machine Learning},
  pages = 	 {22184--22202},
  year = 	 {2022},
  editor = 	 {Chaudhuri, Kamalika and Jegelka, Stefanie and Song, Le and Szepesvari, Csaba and Niu, Gang and Sabato, Sivan},
  volume = 	 {162},
  series = 	 {Proceedings of Machine Learning Research},
  month = 	 {17--23 Jul},
  publisher =    {PMLR},
  url = 	 {https://proceedings.mlr.press/v162/verma22c.html},
}

@inproceedings{madras2018predictresponsiblyimprovingfairness,
  title = {Predict Responsibly: {{Improving}} Fairness and Accuracy by Learning to Defer},
  booktitle = {Advances in Neural Information Processing Systems},
  author = {Madras, David and Pitassi, Toni and Zemel, Richard},
  editor = {Bengio, S. and Wallach, H. and Larochelle, H. and Grauman, K. and {Cesa-Bianchi}, N. and Garnett, R.},
  year = 2018,
  volume = {31},
  publisher = {Curran Associates, Inc.}
}

@inproceedings{mozannar2023predictexactalgorithmslearning,
  title     = {Who Should Predict? Exact Algorithms For Learning to Defer to Humans},
  author    = {Mozannar, Hussein and Lang, Hunter and Wei, Dennis and Sattigeri, Prasanna and Das, Subhro and Sontag, David},
  booktitle = {Proceedings of the 26th International Conference on Artificial Intelligence and Statistics},
  editor    = {Ruiz, Francisco and Dy, Jennifer and van de Meent, Jan-Willem},
  series    = {Proceedings of Machine Learning Research},
  volume    = {206},
  pages     = {10520--10545},
  publisher = {PMLR},
  year      = {2023},
  url       = {https://proceedings.mlr.press/v206/mozannar23a.html}
}

@incollection{TURVEY2012495,
title = {Jury Psychology},
editor = {V.S. Ramachandran},
booktitle = {Encyclopedia of Human Behavior (Second Edition)},
publisher = {Academic Press},
edition = {Second},
address = {San Diego},
pages = {495-502},
year = {2012},
isbn = {978-0-08-096180-4},
doi = {https://doi.org/10.1016/B978-0-12-375000-6.00216-0},
url = {https://www.sciencedirect.com/science/article/pii/B9780123750006002160},
author = {B.E. Turvey and J.L. Freeman},
}

@article{Kovic04032019,
author = {Marko Kovic and Silje Kristiansen},
title = {The gambler’s fallacy fallacy (fallacy)},
journal = {Journal of Risk Research},
volume = {22},
number = {3},
pages = {291--302},
year = {2019},
publisher = {Routledge},
doi = {10.1080/13669877.2017.1378248},
URL = {https://doi.org/10.1080/13669877.2017.1378248},
eprint = { https://doi.org/10.1080/13669877.2017.1378248}
}

@InProceedings{tailor2024learningdeferpopulationmetalearning,
  title = 	 {Learning to Defer to a Population: A Meta-Learning Approach},
  author =       {Tailor, Dharmesh and Patra, Aditya and Verma, Rajeev and Manggala, Putra and Nalisnick, Eric},
  booktitle = 	 {Proceedings of The 27th International Conference on Artificial Intelligence and Statistics},
  pages = 	 {3475--3483},
  year = 	 {2024},
  editor = 	 {Dasgupta, Sanjoy and Mandt, Stephan and Li, Yingzhen},
  volume = 	 {238},
  series = 	 {Proceedings of Machine Learning Research},
  month = 	 {02--04 May},
  publisher =    {PMLR},
  url = 	 {https://proceedings.mlr.press/v238/tailor24a.html},
}

@inproceedings{hemmer2023learningdeferlimitedexpert,
  title     = {Learning to Defer with Limited Expert Predictions},
  author    = {Hemmer, Patrick and Thede, Lukas and V{\"o}ssing, Michael and Jakubik, Johannes and K{\"u}hl, Niklas},
  booktitle = {Proceedings of the AAAI Conference on Artificial Intelligence},
  volume    = {37},
  number    = {5},
  pages     = {6002--6011},
  year      = {2023},
  doi       = {10.1609/aaai.v37i5.25742}
}

@InProceedings{charusaie2022sampleefficientlearningpredictors,
  title = 	 {Sample Efficient Learning of Predictors that Complement Humans},
  author =       {Charusaie, Mohammad-Amin and Mozannar, Hussein and Sontag, David and Samadi, Samira},
  booktitle = 	 {Proceedings of the 39th International Conference on Machine Learning},
  pages = 	 {2972--3005},
  year = 	 {2022},
  editor = 	 {Chaudhuri, Kamalika and Jegelka, Stefanie and Song, Le and Szepesvari, Csaba and Niu, Gang and Sabato, Sivan},
  volume = 	 {162},
  series = 	 {Proceedings of Machine Learning Research},
  month = 	 {17--23 Jul},
  publisher =    {PMLR},
  url = 	 {https://proceedings.mlr.press/v162/charusaie22a.html},
}

@ARTICLE{chow-decision-functions,
  author={Chow, C. K.},
  journal={IRE Transactions on Electronic Computers}, 
  title={An optimum character recognition system using decision functions}, 
  year={1957},
  volume={EC-6},
  number={4},
  pages={247-254},
  doi={10.1109/TEC.1957.5222035}}

@inproceedings{NEURIPS2022_bc8f76d9,
 author = {Narasimhan, Harikrishna and Jitkrittum, Wittawat and Menon, Aditya K and Rawat, Ankit and Kumar, Sanjiv},
 booktitle = {Advances in Neural Information Processing Systems},
 editor = {S. Koyejo and S. Mohamed and A. Agarwal and D. Belgrave and K. Cho and A. Oh},
 pages = {29292--29304},
 publisher = {Curran Associates, Inc.},
 title = {Post-hoc estimators for learning to defer to an expert},
 url = {https://proceedings.neurips.cc/paper_files/paper/2022/file/bc8f76d9caadd48f77025b1c889d2e2d-Paper-Conference.pdf},
 volume = {35},
 year = {2022}
}

@inproceedings{Kamar2016HybridIntelligence,
  author    = {Ece Kamar},
  title     = {Directions in Hybrid Intelligence: Complementing AI Systems with Human Intelligence},
  booktitle = {Proceedings of the Twenty-Fifth International Joint Conference on Artificial Intelligence (IJCAI-16)},
  year      = {2016},
  url       = {https://www.microsoft.com/en-us/research/wp-content/uploads/2016/02/HybridIntelligence.pdf}
}

@article{Dellermann2019HybridIntelligence,
  author  = {Dominik Dellermann and Philipp Ebel and Matthias S{\"o}llner and Jan Marco Leimeister},
  title   = {Hybrid Intelligence},
  journal = {Business \& Information Systems Engineering},
  year    = {2019},
  volume  = {61},
  number  = {5},
  pages   = {637--643},
  doi     = {10.1007/s12599-019-00595-2},
  url     = {https://link.springer.com/article/10.1007/s12599-019-00595-2}
}

@article{Akata2020ResearchAgendaHI,
  author  = {Zeynep Akata and Dan Balliet and Maarten de Rijke and Frank Dignum and Virginia Dignum and A. E. Eiben and Antske Fokkens and Davide Grossi and Koen V. Hindriks and Holger Hoos and Haley Hung and Catholijn Jonker and Christof Monz and Mark Neerincx and Frans Oliehoek and Henry Prakken and Stefan Schlobach and Linda C. van der Gaag and Frank van Harmelen and Herke van Hoof and M. Birna van Riemsdijk and Aimee van Wynsberghe and Rineke Verbrugge and Bart Verheij and Paul Vossen and Max Welling},
  title   = {A Research Agenda for Hybrid Intelligence: Augmenting Human Intellect With Collaborative, Adaptive, Responsible, and Explainable Artificial Intelligence},
  journal = {Computer},
  year    = {2020},
  volume  = {53},
  number  = {8},
  pages   = {18--28},
  doi     = {10.1109/MC.2020.2996587},
  url     = {https://www.computer.org/csdl/magazine/co/2020/08/09153877/1lUB5gL2CnS}
}

@article{bruni2012factors,
  title={Factors involved in discrepant preliminary radiology resident interpretations of neuroradiological imaging studies: a retrospective analysis},
  author={Bruni, Silvio G and Bartlett, Eric and Yu, Eugene},
  journal={American Journal of Roentgenology},
  volume={198},
  number={6},
  pages={1367--1374},
  year={2012},
  publisher={American Roentgen Ray Society}
}

@article{Figalova2024FatigueUnderloadL3,
  title   = {Fatigue and mental underload further pronounced in L3 conditionally automated driving: Results from an EEG experiment on a test track},
  author  = {Figalová, Nikol and Bieg, Hans Joachim and Schulz, Michael and Pichen, Jürgen and Baumann, Martin and Chuang, Lewis and Pollatos, Olga},
  journal = {ArXiv preprint arXiv:2405.18114},
  year    = {2024}
}

@article{Peukert2023ATCFatigueDynamics,
  title   = {Subjective and objective fatigue dynamics in air traffic control},
  author  = {Peukert, M. and others},
  journal = {Journal of Sleep Research},
  year    = {2023}
}

@article{Legler2025RepetitionBias,
  author       = {Eric Legler and Dar{\'\i}o Cuevas Rivera and Sarah Schw{\"o}bel and Ben J. Wagner and Stefan Kiebel},
  title        = {Cognitive computational model reveals repetition bias in a sequential decision-making task},
  journal      = {Communications Psychology},
  year         = {2025},
  volume       = {3},
  number       = {1},
  pages        = {92},
  doi          = {10.1038/s44271-025-00271-0},
  pmid         = {40514401},
  pmcid        = {PMC12166051}
}

@article{Urai2019ChoiceHistory,
  author  = {Anne E. Urai and Jan Willem de Gee and Konstantinos Tsetsos and Tobias H. Donner},
  title   = {Choice history biases subsequent evidence accumulation},
  journal = {eLife},
  year    = {2019},
  volume  = {8},
  pages   = {e46331},
  doi     = {10.7554/eLife.46331},
  pmid    = {31264959},
  pmcid   = {PMC6606080}
}

@article{vandeWouw2024BiasedExpectations,
  author  = {Didrika S. van de Wouw and Ryan T. McKay and Nicholas Furl},
  title   = {Biased expectations about future choice options predict sequential economic decisions},
  journal = {Communications Psychology},
  year    = {2024},
  volume  = {2},
  number  = {1},
  pages   = {119},
  doi     = {10.1038/s44271-024-00172-8},
  pmid    = {39695326},
  pmcid   = {PMC11655870}
}

@inproceedings{verma2023learning,
    title={{Learning to Defer to Multiple Experts: Consistent Surrogate Losses, Confidence Calibration, and Conformal Ensembles}},
    author={Verma, Rajeev and Barrej{\'o}n, Daniel and Nalisnick, Eric},
    booktitle={Proceedings of the 26th International Conference on Artificial Intelligence and Statistics},
    year={2023}
}

@InProceedings{mozannar2021consistentestimatorslearningdefer,
  title = 	 {Consistent Estimators for Learning to Defer to an Expert},
  author =       {Mozannar, Hussein and Sontag, David},
  booktitle = 	 {Proceedings of the 37th International Conference on Machine Learning},
  pages = 	 {7076--7087},
  year = 	 {2020},
  editor = 	 {III, Hal Daumé and Singh, Aarti},
  volume = 	 {119},
  series = 	 {Proceedings of Machine Learning Research},
  month = 	 {13--18 Jul},
  publisher =    {PMLR},
  url = 	 {https://proceedings.mlr.press/v119/mozannar20b.html}
}

@article{Pan2022PilotsFatiguefNIRS,
  title   = {Research on the Identification of Pilots' Fatigue Status Based on Functional Near-Infrared Spectroscopy},
  author  = {Pan, Ting and Wang, Haibo and Si, Haiqing and Liu, Haibo and Xu, Mengyue},
  journal = {Aerospace},
  volume  = {9},
  number  = {3},
  pages   = {173},
  year    = {2022},
  doi     = {10.3390/aerospace9030173}
}

@TECHREPORT{Krizhevsky09learningmultiple,
  author       = {Alex Krizhevsky},
  title        = {Learning Multiple Layers of Features from Tiny Images},
  institution  = {University of Toronto},
  year         = {2009},
  url          = {https://www.cs.toronto.edu/~kriz/learning-features-2009-TR.pdf}
}

@INPROCEEDINGS {peterson2019humanuncertaintymakesclassification,
author = { Peterson, Joshua and Battleday, Ruairidh and Griffiths, Thomas and Russakovsky, Olga },
booktitle = { 2019 IEEE/CVF International Conference on Computer Vision (ICCV) },
title = {{ Human Uncertainty Makes Classification More Robust}},
year = {2019},
volume = {},
ISSN = {},
pages = {9616-9625},
doi = {10.1109/ICCV.2019.00971},
url = {https://doi.ieeecomputersociety.org/10.1109/ICCV.2019.00971},
publisher = {IEEE Computer Society},
address = {Los Alamitos, CA, USA},
month =Nov}

@inproceedings{irvin2019chexpertlargechestradiograph,
  title     = {{CheXpert}: A Large Chest Radiograph Dataset with Uncertainty Labels and Expert Comparison},
  author    = {Irvin, Jeremy and Rajpurkar, Pranav and Ko, Michael and Yu, Yifan and Ciurea-Ilcus, Silviana and Chute, Chris and Marklund, Henrik and Haghgoo, Behzad and Ball, Robyn and Shpanskaya, Katie and Seekins, Jayne and Mong, David A. and Halabi, Safwan S. and Sandberg, Jesse K. and Jones, Ricky and Larson, David B. and Langlotz, Curtis P. and Patel, Bhavik N. and Lungren, Matthew P. and Ng, Andrew Y.},
  booktitle = {Proceedings of the AAAI Conference on Artificial Intelligence},
  volume    = {33},
  number    = {01},
  pages     = {590--597},
  year      = {2019},
  doi       = {10.1609/aaai.v33i01.3301590}
}

@inproceedings{davidson2017automatedhatespeechdetection,
  title     = {Automated Hate Speech Detection and the Problem of Offensive Language},
  author    = {Davidson, Thomas and Warmsley, Dana and Macy, Michael and Weber, Ingmar},
  booktitle = {Proceedings of the International AAAI Conference on Web and Social Media},
  volume    = {11},
  number    = {1},
  pages     = {512--515},
  year      = {2017},
  doi       = {10.1609/icwsm.v11i1.14955}
}

@article{vosshenrich2021quantifying,
  title   = {Quantifying Radiology Resident Fatigue: Analysis of Preliminary Reports},
  author  = {Vosshenrich, Jan and Brantner, Philipp and Cyriac, Joshy and Boll, Daniel T. and Merkle, Elmar M. and Heye, Tobias},
  journal = {Radiology},
  year    = {2021},
  volume  = {298},
  number  = {3},
  pages   = {632--639},
  doi     = {10.1148/radiol.2021203486},
  url     = {https://doi.org/10.1148/radiol.2021203486}
}

@inproceedings{kern-etal-2023-annotation,
    title = "Annotation Sensitivity: Training Data Collection Methods Affect Model Performance",
    author = "Kern, Christoph  and
      Eckman, Stephanie  and
      Beck, Jacob  and
      Chew, Rob  and
      Ma, Bolei  and
      Kreuter, Frauke",
    editor = "Bouamor, Houda  and
      Pino, Juan  and
      Bali, Kalika",
    booktitle = "Findings of the Association for Computational Linguistics: EMNLP 2023",
    month = dec,
    year = "2023",
    address = "Singapore",
    publisher = "Association for Computational Linguistics",
    url = "https://aclanthology.org/2023.findings-emnlp.992/",
    doi = "10.18653/v1/2023.findings-emnlp.992",
    pages = "14874--14886",
}

@inproceedings{abercrombie-etal-2023-temporal,
    title = "Temporal and Second Language Influence on Intra-Annotator Agreement and Stability in Hate Speech Labelling",
    author = "Abercrombie, Gavin  and
      Hovy, Dirk  and
      Prabhakaran, Vinodkumar",
    editor = "Prange, Jakob  and
      Friedrich, Annemarie",
    booktitle = "Proceedings of the 17th Linguistic Annotation Workshop (LAW-XVII)",
    month = jul,
    year = "2023",
    address = "Toronto, Canada",
    publisher = "Association for Computational Linguistics",
    url = "https://aclanthology.org/2023.law-1.10/",
    doi = "10.18653/v1/2023.law-1.10",
    pages = "96--103",
}

@misc{zhang2026fatigueawarelearningdeferconstrained,
      title={Fatigue-Aware Learning to Defer via Constrained Optimisation}, 
      author={Zheng Zhang and Cuong C. Nguyen and David Rosewarne and Kevin Wells and Gustavo Carneiro},
      year={2026},
      eprint={2604.00904},
      archivePrefix={arXiv},
      primaryClass={cs.LG},
      url={https://arxiv.org/abs/2604.00904}, 
}

@InCollection{Newell81,
  author =       "A. Newell and P. S. Rosenbloom",
  title =        "Mechanisms of Skill Acquisition and the Law of
                  Practice", 
  booktitle =    "Cognitive Skills and Their Acquisition",
  pages =        "1--51",
  publisher =    "Lawrence Erlbaum Associates, Inc.",
  year =         "1981",
  editor =       "J. R. Anderson",
  chapter =      "1",
  address =      "Hillsdale, NJ"
}

@article{alshabibi2020impact,
  title   = {Impact of Time of Day on Radiology Image Interpretations},
  author  = {Alshabibi, Abdulaziz S. and Suleiman, Moayyad E. and Tapia, Kriscia A. and Heard, Robert and Brennan, Patrick C.},
  journal = {Clinical Radiology},
  volume  = {75},
  number  = {10},
  pages   = {746--756},
  year    = {2020},
  doi     = {10.1016/j.crad.2020.05.004},
  pmid    = {32576366}
}

@article{krupinski2010long,
  title   = {Long Radiology Workdays Reduce Detection and Accommodation Accuracy},
  author  = {Krupinski, Elizabeth A. and Berbaum, Kevin S. and Caldwell, Robert T. and Schartz, Kevin M. and Kim, John},
  journal = {Journal of the American College of Radiology},
  volume  = {7},
  number  = {9},
  pages   = {698--704},
  year    = {2010},
  doi     = {10.1016/j.jacr.2010.03.004}
}

@inproceedings{duy2026online,
  title     = {Online Learning-to-Defer with Varying Experts},
  author    = {Montreuil, Yannis and Dang, Hoang Duy and Meyer, Maxime and Ng, Lai Xing and Carlier, Axel and Ooi, Wei Tsang},
  booktitle = {Proceedings of the 29th International Conference on Artificial Intelligence and Statistics},
  series    = {Proceedings of Machine Learning Research},
  volume    = {300},
  pages     = {2701--2709},
  publisher = {PMLR},
  year      = {2026},
  url       = {https://proceedings.mlr.press/v300/montreuil26b.html}
}
\bibliographystyle{tmlr}

\newpage
\appendix
\section{Theoretical Results}

\subsection{Bayes Rule under Expert Drift}
\label{bayes-rule-proof6226}

We first characterize the Bayes-optimal decision rule when the expert's
reliability changes over time. Let $\mathcal{X}$ and $\mathcal{Y}$ denote
the input and label spaces, and assume that the expert also produces a
prediction $m_t\in\mathcal{Y}$ at each time step $t\in\mathcal{T}$. The
label distribution $\mathbb{P}(y\mid x)$ is assumed to be stationary,
whereas the expert's behavior may depend on time.

A classifier $h:\mathcal{X}\rightarrow\mathcal{Y}$ is paired with a
time-dependent rejector
$r:\mathcal{X}\times\mathcal{T}\rightarrow\{0,1\}$. The resulting system
decision is
\[
\delta(x,t)
=
\begin{cases}
h(x), & r(x,t)=0,\\
\bot, & r(x,t)=1,
\end{cases}
\]
where $\bot$ denotes deferral to the expert.

\paragraph{Learning-to-defer risk.}
For an example with label $y$ and expert prediction $m_t$, the system
incurs the cost
\begin{equation}
c_{\mathrm{L2D}}\bigl(\delta(x,t),y,m_t\bigr)
=
\begin{cases}
\mathbb{I}[\hat y\neq y],
    & \delta(x,t)=\hat y\in\mathcal{Y},\\
\mathbb{I}[m_t\neq y],
    & \delta(x,t)=\bot.
\end{cases}
\label{eq:cost-l2d}
\end{equation}
Thus, a model prediction is penalized when the classifier is incorrect,
whereas a deferred prediction is penalized when the expert is incorrect.

For a fixed input $x$ and time step $t$, define
\[
\eta_y(x):=\mathbb{P}(Y=y\mid X=x)
\]
and
\[
\alpha_t(x)
:=
\mathbb{P}(M_t=Y\mid X=x,t),
\]
where $\alpha_t(x)$ is the expert's conditional probability of being
correct at time $t$. Importantly, this definition does not require the
expert prediction and the true label to be conditionally independent.

If the system predicts a class $\hat y$, its conditional risk is
\begin{equation}
\mathcal{R}(x,t,\hat y)
=
\mathbb{P}(Y\neq\hat y\mid X=x)
=
1-\eta_{\hat y}(x).
\label{eq:risk-predict}
\end{equation}
The best model prediction is therefore
\begin{equation}
h^*(x)
=
\arg\max_{y\in\mathcal{Y}}\eta_y(x),
\label{eq:bayes-classifier}
\end{equation}
with conditional risk
\[
1-\max_{y\in\mathcal{Y}}\eta_y(x).
\]

If the system instead defers, its conditional risk is the expert's
probability of error:
\begin{equation}
\mathcal{R}(x,t,\bot)
=
\mathbb{P}(M_t\neq Y\mid X=x,t)
=
1-\alpha_t(x).
\label{eq:risk-defer}
\end{equation}

Comparing these two risks gives the Bayes-optimal rejector:
\begin{equation}
r^*(x,t)
=
\mathbb{I}\left[
\alpha_t(x)
\geq
\max_{y\in\mathcal{Y}}\eta_y(x)
\right].
\label{eq:01-corollary}
\end{equation}
Under this tie-breaking convention, the system defers exactly when the
expert's probability of being correct at time $t$ is at least as large
as the classifier's probability of correctly predicting its most likely
class.

This rule extends the stationary result of
\citet{mozannar2021consistentestimatorslearningdefer}. In the stationary
setting, the comparison is made against a fixed expert reliability. Under
expert drift, that quantity becomes $\alpha_t(x)$ and may change over
time.

\subsection{Consistency of the Per-Step Softmax Surrogate}
\label{sec:con-perstep}

We next show that the per-step softmax surrogate in
Equation~\ref{eq:sm_loss_t} recovers the Bayes rule above.

Let
\[
p_y(x,t)
=
\frac{\exp\{g_y(x)\}}
{\exp\{g_\bot^t(x)\}
+\sum_{y'\in\mathcal{Y}}\exp\{g_{y'}(x)\}},
\qquad y\in\mathcal{Y},
\]
and
\[
p_{\bot}(x,t)
=
\frac{\exp\{g_\bot^t(x)\}}
{\exp\{g_\bot^t(x)\}
+\sum_{y'\in\mathcal{Y}}\exp\{g_{y'}(x)\}}.
\]
The per-step surrogate can then be written as
\begin{equation}
\phi_{\mathrm{SM}}^t
=
-\log p_y(x,t)
-
\mathbb{I}[m_t=y]\log p_\bot(x,t).
\label{eq:psi-expanded}
\end{equation}

Conditioning on $(x,t)$ and taking the expectation over the label and
expert prediction gives
\begin{equation}
C_t(p;x)
=
-\sum_{y\in\mathcal{Y}}
\eta_y(x)\log p_y(x,t)
-
\alpha_t(x)\log p_\bot(x,t).
\label{eq:cond-surrogate}
\end{equation}
This is a weighted cross-entropy over the augmented output space
$\mathcal{Y}\cup\{\bot\}$. Its minimizing probabilities are
\begin{equation}
p_y^*(x,t)
=
\frac{\eta_y(x)}{1+\alpha_t(x)},
\qquad
p_\bot^*(x,t)
=
\frac{\alpha_t(x)}{1+\alpha_t(x)}.
\label{eq:optimal-softmax-probs}
\end{equation}
The common denominator does not affect the ordering of the outputs.
Consequently,
\[
\arg\max_{y\in\mathcal{Y}}p_y^*(x,t)
=
\arg\max_{y\in\mathcal{Y}}\eta_y(x),
\]
and
\[
p_\bot^*(x,t)
\geq
\max_{y\in\mathcal{Y}}p_y^*(x,t)
\iff
\alpha_t(x)
\geq
\max_{y\in\mathcal{Y}}\eta_y(x).
\]
The classifier and rejector induced by the softmax surrogate therefore
coincide with the Bayes-optimal rules in
Equations~\ref{eq:bayes-classifier} and~\ref{eq:01-corollary}. Hence, the
per-step softmax surrogate is Bayes-consistent for the drifting-expert
L2D objective.

\subsection{Consistency of the RNN-Based Softmax Surrogate}
\label{sec:con-rnn}

The RNN-based model additionally conditions its deferral decision on
information accumulated from previous time steps. Let $H_t$ denote the
history available to the model before making the decision at time $t$,
including previously observed expert-correctness feedback. Define
\begin{equation}
\alpha_t(x,H_t)
:=
\mathbb{P}(M_t=Y_t\mid X_t=x,H_t).
\label{eq:history-expert-accuracy}
\end{equation}
Under the assumption that the label distribution remains stationary and
that $Y_t$ is conditionally independent of $H_t$ given $X_t$, the
classifier confidence remains $\eta_y(x)=\mathbb{P}(Y_t=y\mid X_t=x)$.
The Bayes-optimal history-aware rejector is therefore
\begin{equation}
r^*(x,H_t)
=
\mathbb{I}\left[
\alpha_t(x,H_t)
\geq
\max_{y\in\mathcal{Y}}\eta_y(x)
\right].
\label{eq:history-bayes-rejector}
\end{equation}

The sequence-level surrogate is the sum of the per-step losses:
\begin{equation}
\phi_{\mathrm{SM}}^{\mathcal{T}}(g)
=
\sum_{t\in\mathcal{T}}
\phi_{\mathrm{SM}}^t
\bigl(g;X_t,Y_t,M_t,H_t\bigr).
\label{eq:rnn-sequence-loss}
\end{equation}
By linearity of expectation, its population risk decomposes as
\begin{equation}
\mathcal{R}_{\mathrm{RNN}}(g)
=
\sum_{t\in\mathcal{T}}
\mathbb{E}_{X_t,H_t}
\left[
C_t(g;X_t,H_t)
\right],
\label{eq:rnn-risk}
\end{equation}
where
\[
C_t(g;x,H_t)
=
\mathbb{E}
\left[
\phi_{\mathrm{SM}}^t
\mid X_t=x,H_t
\right].
\]

For each $(x,H_t)$, this conditional risk has the same weighted
cross-entropy form as Equation~\ref{eq:cond-surrogate}, with
$\alpha_t(x)$ replaced by $\alpha_t(x,H_t)$. Its optimal softmax
probabilities are therefore
\begin{equation}
p_y^*(x,H_t)
=
\frac{\eta_y(x)}
{1+\alpha_t(x,H_t)},
\qquad
p_\bot^*(x,H_t)
=
\frac{\alpha_t(x,H_t)}
{1+\alpha_t(x,H_t)}.
\label{eq:rnn-optimal-probs}
\end{equation}

Assuming that the recurrent model class is rich enough to represent
these conditional probabilities, minimizing the sequence-level
surrogate recovers the history-aware Bayes rule in
Equation~\ref{eq:history-bayes-rejector}. The RNN-based softmax surrogate
is therefore Bayes-consistent under this realizability assumption.

\subsection{Temporal Complementarity under Expert Drift}
\label{sec:complementarity}

The Bayes rule in Equation~\ref{eq:01-corollary} shows that the optimal
deferral decision depends on a comparison between the classifier's
confidence and the expert's current reliability. In particular, the
system should defer at $(x,t)$ when
\[
\alpha_t(x)
\geq
\max_{y\in\mathcal{Y}}\eta_y(x),
\]
where
\[
\eta_y(x)=\mathbb{P}(Y=y\mid X=x)
\]
and
\[
\alpha_t(x)=\mathbb{P}(M_t=Y\mid X=x,t).
\]

Under our assumption that the input and label distribution remains
stationary, the classifier confidence for a given input does not change
with time. The expert reliability, however, may drift. Consequently, the
relative ordering of the two quantities can change over a sequence:
\begin{equation}
\alpha_{t_1}(x)
<
\max_{y\in\mathcal{Y}}\eta_y(x)
\qquad\text{and}\qquad
\alpha_{t_2}(x)
>
\max_{y\in\mathcal{Y}}\eta_y(x)
\label{eq:temporal-crossing}
\end{equation}
for two time steps $t_1$ and $t_2$. We refer to this change in ordering
as a \emph{temporal crossing}.

A temporal crossing changes the Bayes-optimal routing decision for the
same input: the classifier is preferred at $t_1$, whereas deferral to
the expert is preferred at $t_2$. A model that treats expert reliability
as stationary cannot represent this change using the expert's temporal
state alone. In contrast, a time-aware deferral model can adjust its
decision as the expert moves from one side of the classifier-confidence
threshold to the other.

The benefit of modeling expert drift therefore arises not merely because
the expert's accuracy changes, but because these changes can alter which
component is more reliable. If the expert remains uniformly better or
uniformly worse than the classifier throughout the sequence, temporal
adaptation provides no additional routing opportunity. When temporal
crossings occur, however, tracking the expert's current reliability
allows the system to defer during periods when the expert is more
reliable and rely more heavily on the classifier when the expert's
reliability declines.

EDA-L2D is designed to exploit this temporal structure. Its recurrent
deferral component uses the history of the sequence and the available
expert-correctness feedback to estimate the expert's current state and
adapt the deferral decision accordingly. The model need not observe
$\alpha_t(x)$ directly; rather, it learns temporal patterns that help it track when the expert is likely to cross the classifier's reliability threshold.
\section{Expert Simulation}
\label{app:expert-simulation}

Training an L2D model requires the expert's answer for every training input. Existing datasets do not provide such data: \textsc{CIFAR-10H} collects human annotations only on the \textsc{CIFAR-10} test set, and CheXpert and the hate speech dataset contain no answers of individual experts over time. We therefore simulate the expert, following the setting of Section~\ref{sec:expert-setting}. The task $\mathbb{P}(\ry \mid x)$ is stationary, and only the expert's predictions $\mathbb{P}_t(\rsm_t \mid x)$ change with $t$. We capture this change through the expert's accuracy over time and use a surrogate expert whose probability of being correct depends only on the time step:
\[
\mathbb{P}_t(\rsm_t = \ry \mid x) = p_t ,
\]
where $p_t$ is a drift curve. In step $t$, the surrogate expert returns the true label with probability $p_t$ (a Bernoulli draw) and the incorrect label otherwise. This yields expert answers for the training, validation and test sets, and a known curve that we can modify at test time for the robustness tests (Appendix~\ref{app:robustness}).

\subsection{Drifting Annotators in CIFAR-10H}
\label{app:cifar10h-annotators}
On \textsc{CIFAR-10}, $p_t$ is estimated from real annotators in \textsc{CIFAR-10H}~\citep{peterson2019humanuncertaintymakesclassification} .  \textsc{CIFAR-10H} provides human labels for all 10{,}000 images of the \textsc{CIFAR-10} test set, collected from 2{,}571 crowd workers, with about 51 labels per image. Each annotator labels 200 images in sequence, interleaved with 10 attention checks. The raw data record each annotator's answers in presentation order, together with response times. This makes it possible to follow an individual annotator's accuracy over the course of a session, which is the drift we want to simulate. Human accuracy is high on average (94.9\%), but it varies across annotators and, for some annotators, changes considerably within a session.

\begin{figure}[t]
\centering
\includegraphics[width=0.8\textwidth]{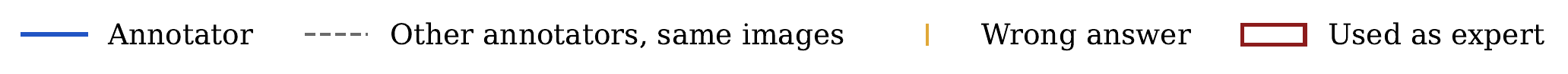}\\[4pt]
\begin{minipage}[t]{0.48\textwidth}
  \centering
  \includegraphics[width=\linewidth]{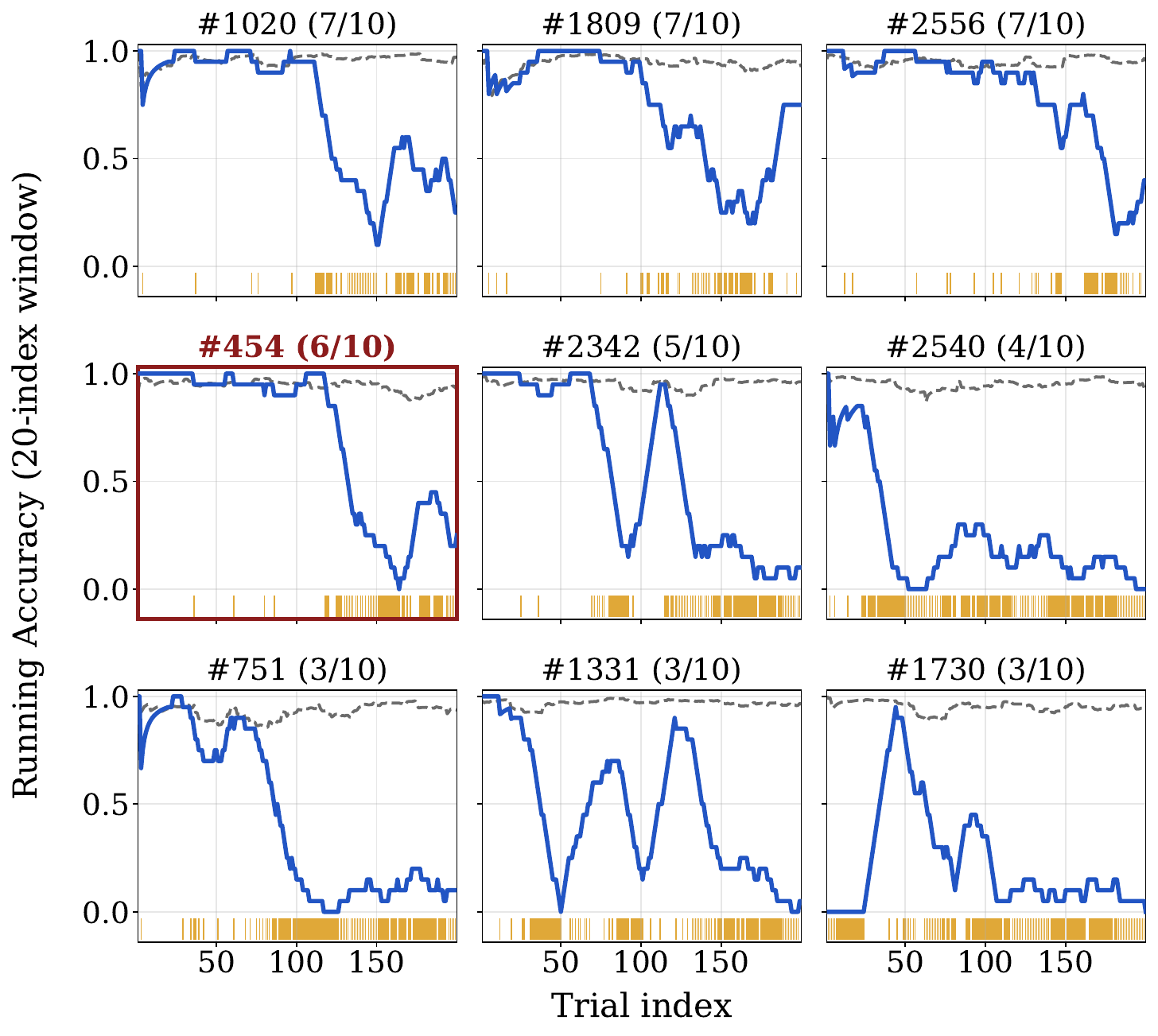}\\
  {\small (a) Below the attention-check threshold}
\end{minipage}\hfill
\begin{minipage}[t]{0.48\textwidth}
  \centering
  \includegraphics[width=\linewidth]{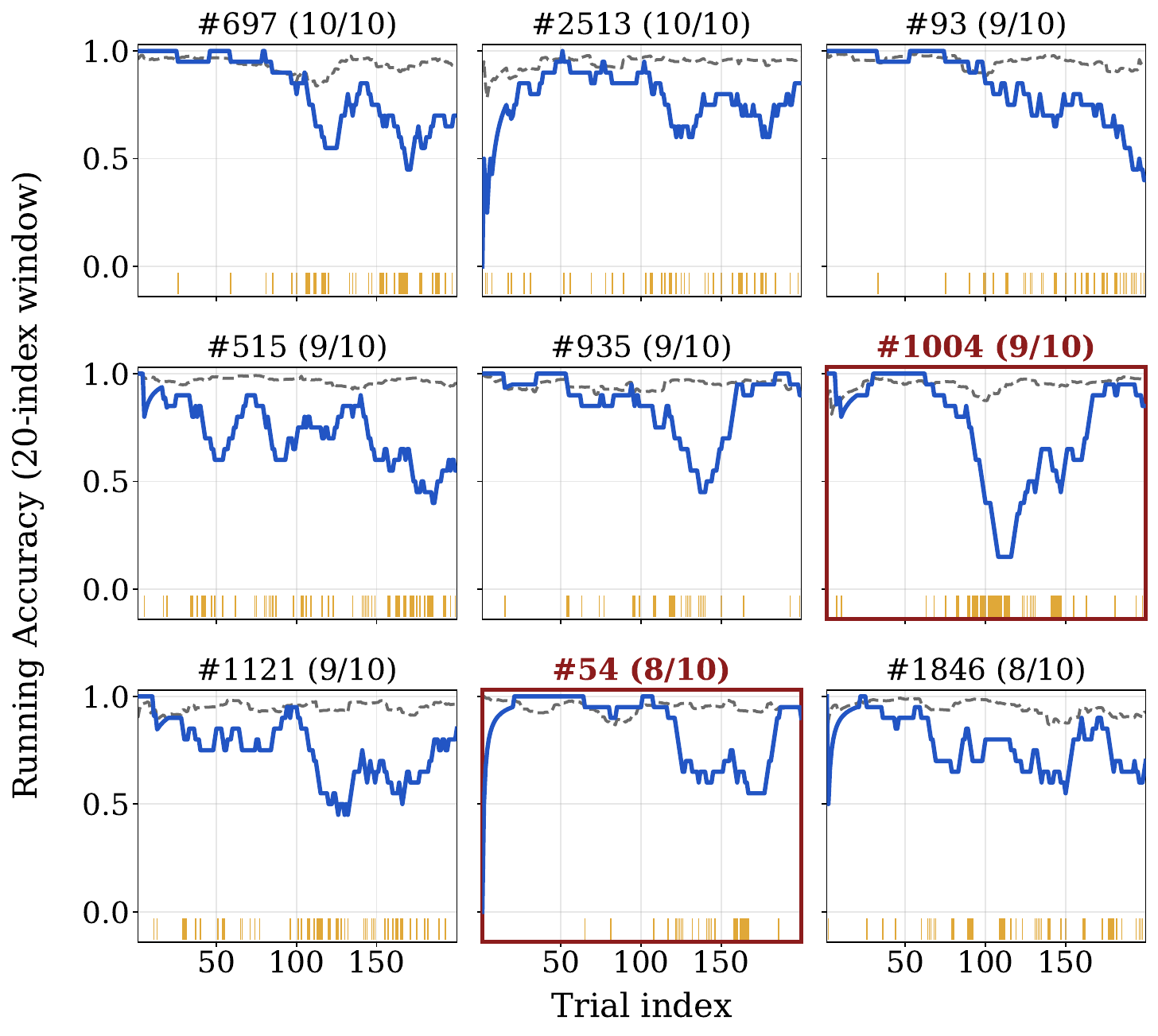}\\
  {\small (b) Above the attention-check threshold}
\end{minipage}
\caption{The 18 \textsc{CIFAR-10H} annotators whose accuracy drops by more than 10 percentage points from the first to the second half of their 200 trials. (a) Annotators below the 75\% attention-check threshold; (b) annotators above it. Titles give the annotator id and the number of attention checks passed. Curves are running accuracies over 20 trials. The other annotators stay accurate on the same images, so the drops are not caused by harder images.}

\label{fig:cifar10h-annotators}
\end{figure}

\paragraph{Annotator selection.}
\label{app:annotator-selection}
We examine whether individual annotators in CIFAR-10H exhibit systematic changes in reliability over the course of an annotation session. For each of the 2,571 annotators, we consider the 200 non-attention-check trials in presentation order and divide them into two consecutive halves of 100 trials. We compute the change in mean accuracy from the first to the second half and identify 18 annotators whose accuracy decreases by more than 10 percentage points. Figure~\ref{fig:cifar10h-annotators} shows all 18 annotators, grouped by whether they pass the 75\% attention-check threshold used by \citet{peterson2019humanuncertaintymakesclassification}; the three annotators we use as experts are framed. Failing the attention checks is itself a form of drift that we wish to model, so we do not exclude these annotators a priori. However, annotators with very low attention scores remain close to chance accuracy ($0.1$ for ten classes) for most of their session (e.g.\ \#751, \#1730, \#2540). This pattern is more consistent with disengaged or random responding than with a change in reliability, and an expert who is almost always wrong offers little to learn for deferral. We therefore select three annotators whose curves combine long periods of high accuracy with distinct, structured forms of drift:
\begin{itemize}
  \item \textbf{\#454} (6/10 attention checks) is nearly perfect for the first $\sim$120 trials and then deteriorates sharply, which is characteristic of an annotator who is attentive early in the session and disengages later.
  \item \textbf{\#1004} (9/10) shows a transient drop to about $0.15$ around trials 100--120, followed by a recovery to its initial level, consistent with a temporary lapse.
  \item \textbf{\#54} (8/10) improves over the first trials, stays near-perfect for most of the session, and declines in the second half before partially recovering.
\end{itemize}
The latter two pass the attention checks and change more gradually, and so represent the more realistic case of a diligent annotator whose reliability varies over the session; \#454 represents a more extreme, abrupt change.

\subsection{Temporal Variation in Radiologist Performance}
\label{app:chexpert-drift}

We further examine evidence of temporal variation in expert performance in a clinical setting. \citet{vosshenrich2021quantifying} retrospectively analyzed 117,402 radiology reports written by residents between September 2017 and March 2020. For each study, they compared the resident's preliminary report with the corresponding staff-reviewed final report using the Jaccard similarity coefficient.

Figure~4 of \citet{vosshenrich2021quantifying} reports mean report similarity over five two-hour intervals for each weekday for the general resident cohort, while Figure~3b reports the corresponding trajectory for the emergency-department cohort at a finer hourly resolution, with ten measurements per weekday. Both cohorts exhibit systematic temporal variation within individual workdays and across the workweek. Report similarity generally decreases over the course of a workday, with temporary midday increases suggesting partial recovery following lunch breaks; this recovery becomes less pronounced toward the end of the workweek.

Because report similarity is not a direct measure of diagnostic accuracy, we use it only to characterize the temporal pattern of expert drift. We digitize the mean report-similarity trajectories from the two figures (25 points for the general cohort, 50 for the emergency-department cohort), double each value's deviation from the trajectory mean and add small uniform jitter, which makes the within- and across-day fluctuations more pronounced while keeping their structure.

We then rescale each trajectory linearly to the accuracy range $[0.695, 0.885]$,
\[
r_t = 0.695 + (0.885 - 0.695)\,\frac{s_t - s_{\min}}{s_{\max} - s_{\min}} .
\]
This range is the span of the balanced accuracy, $(\mathrm{TPR} + 1 - \mathrm{FPR})/2$, of the three individual radiologists on the five competition pathologies in Figure~3 of \citet{irvin2019chexpertlargechestradiograph}. It runs from $0.695$ (Consolidation; $\mathrm{FPR}=0.09$, $\mathrm{TPR}=0.48$) to $0.885$ (Pleural Effusion; $\mathrm{FPR}=0.05$, $\mathrm{TPR}=0.82$). We use individual radiologists rather than their majority vote because in deployment each case is read by a single expert, and balanced accuracy because it accounts for both detecting and ruling out a finding.

\begin{figure}[!htbp]
  \centering
  \includegraphics[width=0.6\linewidth]{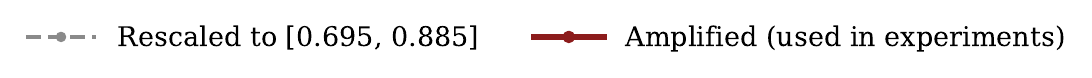}\\[-2pt]
  \begin{subfigure}[b]{0.48\linewidth}
    \centering
    \includegraphics[width=\linewidth]{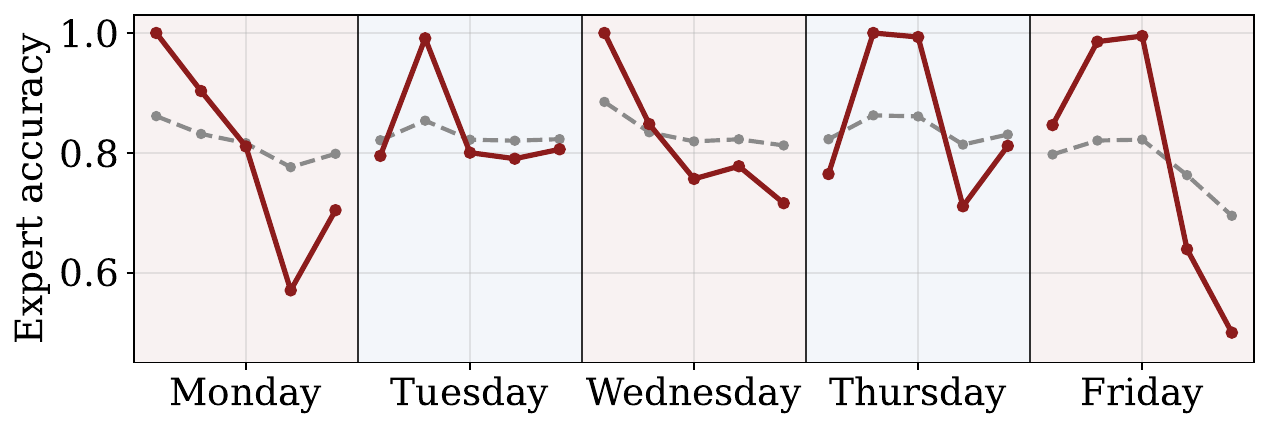}
    \caption{General cohort ($\times 2$, $\times 6$).}
    \label{fig:chexpert-drift-curves-general}
  \end{subfigure}
  \hfill
  \begin{subfigure}[b]{0.48\linewidth}
    \centering
    \includegraphics[width=\linewidth]{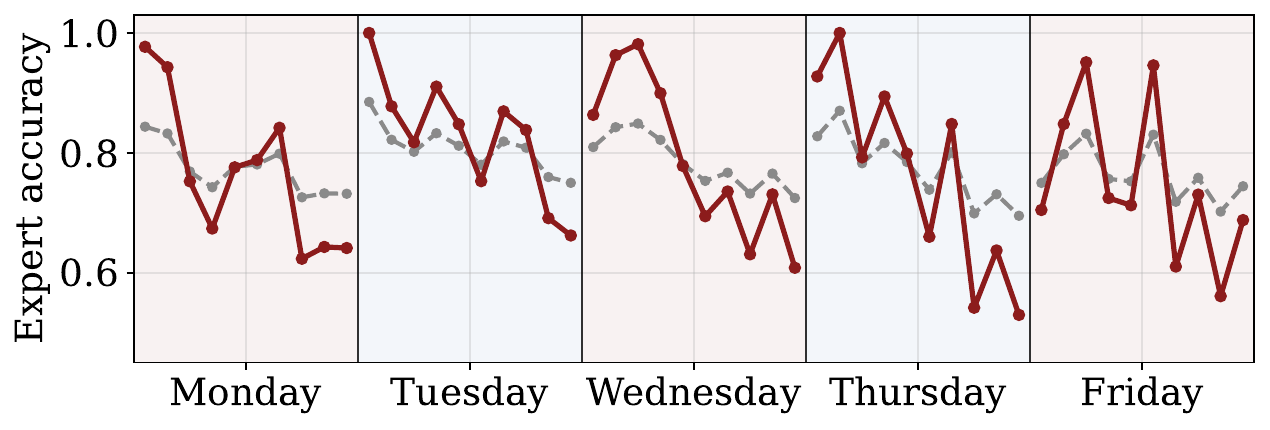}
    \caption{Emergency-department cohort ($\times 2$, $\times 3$).}
    \label{fig:chexpert-drift-curves-emergency}
  \end{subfigure}
  \caption{Expert-accuracy trajectories used for the CheXpert experiments, digitized from Figures~4 and~3b of \citet{vosshenrich2021quantifying}. The dashed line in each panel shows the trajectory rescaled to $[0.695, 0.885]$; the solid line shows it after amplifying the differences between weekdays and the fluctuations within each day by the factors given below each panel, then clipping to $[0.5, 1]$.}
  \label{fig:chexpert-drift-curves}
\end{figure}

Within this range the drift is too small to change the deferral decision over the week. We therefore amplify it around the weekly mean $\bar r$, separately across and within days:
\[
a_t = \operatorname{clip}\!\Big(\bar r + 2\,\big(\bar r_{d(t)} - \bar r\big) + K_D\,\big(r_t - \bar r_{d(t)}\big),\; 0.5,\; 1\Big),
\]
where $\bar r_d$ is the mean over weekday $d$. We use $K_D = 6$ for the general cohort and $K_D = 3$ for the emergency-department cohort, whose within-day deviations are already larger; with $K_D = 6$, most of its days would be clipped into a two-level curve. The resulting trajectories have mean $0.82$ and range $[0.5, 1]$ and mean $0.78$ and range $[0.53, 1]$. They exceed the accuracies observed for the CheXpert radiologists and should be read as a stress test of how well each method follows a strongly drifting expert, not as an estimate of real radiologist performance. In short, the fatigue study determines the temporal structure of the drift, and the CheXpert radiologists determine its accuracy scale before amplification (Figures~\ref{fig:chexpert-drift-curves-general} and~\ref{fig:chexpert-drift-curves-emergency}).\\

\subsection{Synthetic Expert Curves for Hate Speech}
\label{app:hs-curves}
The expert drift used for CheXpert and CIFAR-10 above is grounded in evidence about real annotators. For hate speech, we have no comparable record of a single expert whose accuracy is measured over a controlled sequence. We therefore complement that evidence with three simulated accuracy curves, each motivated by a mechanism reported in the literature:

\begin{itemize}
    \item \textbf{Linear decline.}
    Studies of abusive-language annotation show that the judgments of the
    same annotator can shift between labeling sessions
    \citep{abercrombie-etal-2023-temporal} and that annotations are sensitive
    to the conditions under which they are collected
    \citep{kern-etal-2023-annotation}. A steady decline provides the simplest
    representation of this type of drift.

    \item \textbf{Warm-up followed by fatigue.}
    This curve is motivated by the workload-dependent performance model of
    \citet{zhang2026fatigueawarelearningdeferconstrained}, in which performance
    initially improves as a person settles into the task and subsequently
    declines as fatigue accumulates. It is also consistent with the vigilance
    decrement observed in prolonged monitoring tasks.

    \item \textbf{Practice curve.}
    This curve is motivated by the power law of practice \citep{Newell81},
    under which performance improves with experience but with diminishing
    gains. It allows us to evaluate drift in the direction opposite to the
    linear-decline setting.
\end{itemize}

These curves are stylized test cases and are not fitted to any individual annotator. Their exact forms and parameters are provided below. Formally, at step $t \in \{1,\dots,T\}$ the expert answers the tweet correctly with probability $p_t$, drawn independently for every tweet, and a miss earns no credit, so the accuracy of the expert at step $t$ is $p_t$. We write $u = (t-1)/(T-1) \in [0,1]$ with $T = 10$, so that $u=0$ at the first step and $u=1$ at the last.

\paragraph{Linear Decline.} Accuracy falls at a constant rate from $p_{\mathrm{start}}$ at the first step to $p_{\mathrm{end}}$ at the last step,
\begin{equation}
p_t^{\mathrm{lin}} = p_{\mathrm{start}} - (p_{\mathrm{start}} - p_{\mathrm{end}})\,u,
\end{equation}
with $p_{\mathrm{start}} = 1.0$ and $p_{\mathrm{end}} = 0.55$, i.e.\ $p_t^{\mathrm{lin}} = 1 - 0.45\,u$.

\paragraph{Warm-up then Fatigue.} Following the two-phase performance model of \citet{zhang2026fatigueawarelearningdeferconstrained}, accuracy first rises quadratically from an initial level $w_0$ to a peak $w_{\mathrm{peak}}$ during the warm-up phase, which ends at $\hat{u}$, and then declines along a logistic curve towards a floor $w_{\mathrm{base}}$,
\begin{equation}
p_t^{\mathrm{fat}} =
\begin{cases}
w_0 + (w_{\mathrm{peak}} - w_0)\,\bigl(u/\hat{u}\bigr)^2, & u \le \hat{u}, \\[4pt]
w_{\mathrm{base}} + \dfrac{w_{\mathrm{peak}} - w_{\mathrm{base}}}{1 + \exp\{K\,(u - \bar{u})\}}, & u > \hat{u}.
\end{cases}
\end{equation}
Here $\bar{u}$ is the point at which the decline is halfway complete and $K$ controls how sharp it is. We use $w_0 = 0.85$, $w_{\mathrm{peak}} = 0.99$, $w_{\mathrm{base}} = 0.60$, $\hat{u} = 0.15$, $\bar{u} = 0.55$, and $K = 12$. The two phases join with a gap below $0.005$ at $u = \hat{u}$.

\paragraph{Power-law Practice.} Accuracy improves with experience with diminishing gains,
\begin{equation}
p_t^{\mathrm{pow}} = p_\infty - (p_\infty - p_{\mathrm{start}})\,t^{-c},
\end{equation}
with $p_{\mathrm{start}} = 0.65$, $p_\infty = 1.0$, and $c = 0.8$. This curve is indexed by the step $t$ rather than by $u$; at the first step, $t^{-c}=1$ and $p_1^{\mathrm{pow}} = p_{\mathrm{start}}$.

All parameters are our own choices for a ten-step window and are not fitted to data. Each curve crosses the accuracy of the classifier alone, which is about $0.88$, at least once.
\section{Implementation Details}
\label{app:implementation}

All three variants are trained with our surrogate loss (Equation~\ref{eq:sm_loss}), with $K+1$ outputs where output $K+1$ means defer; the RNN variant uses its sequential form (Equation~\ref{eq:sm_loss_rnn}). We use the same three datasets as \citet{mozannar2021consistentestimatorslearningdefer} and follow their setup. They introduce a weight $\alpha$ on the classification term for inputs on which the expert is correct; we set $\alpha = 1$, which gives the loss exactly as written. All variants face the same simulated expert (Appendix~\ref{app:expert-simulation}): the same drift curve and, within a seed, the same expert answers. The variants differ only in how time enters the model.

\subsection{The Variants and Baseline}
\paragraph{Naive L2D.} A single time-agnostic model is trained on the items of all steps pooled together. It is the standard L2D model and does not receive the time step or any information about the expert's earlier responses.

\paragraph{EDA-L2D, Per-step Param.} The model has the same architecture and loss as Naive L2D, but one defer output per time step instead of a single one. The class outputs and the backbone are shared across steps, and at step $t$ the model uses the class outputs together with the defer output of step $t$. Each defer head can thus follow its own step of the drift, while the classifier is learned from all the data. On hate speech we therefore use separate models.

\paragraph{EDA-L2D, RNN Param.} The backbone of Naive L2D encodes each item, and its features give the $K$ class outputs as in Naive L2D. A recurrent cell (GRU or LSTM) runs over the sequence; its input at step $t$ is the item feature and the expert's correctness at step $t-1$, and its state gives the defer output. The class outputs are thus the same as in Naive L2D, and the decision to defer depends on the expert's history.

\subsection{Common Training Setup}
For each dataset, all variants share the backbone, the classifier head, the optimiser and the weight decay, which we take from the original implementation of \citet{mozannar2021consistentestimatorslearningdefer}.
Naive L2D and EDA-L2D, Per-step param.\ have the same classifier as their L2D model and differ only in the defer output. We therefore start from their settings but still select the learning rate on the validation set for every dataset. EDA-L2D, RNN param.\ adds a recurrent module that has no counterpart in their setup, so we select both its learning rate and its architecture on the validation set: the recurrent cell, the hidden size and the number of layers.

On every dataset, each batch consists of whole sequences in time order, so all variants see the same number of inputs per batch, and EDA-L2D, RNN param.\ can run over each sequence within a batch. Naive L2D and EDA-L2D, Per-step param.\ treat the inputs of a batch independently.

\subsection{CIFAR-10 Setup}
\label{app:imple-cifar10}

\subsubsection{Temporal Rescaling}
\label{app:rescaling}
For each selected annotator (Appendix~\ref{app:cifar10h-annotators}), the raw CIFAR-10H response sequence consists of the same 200 non-attention-check trials used for the selection, each recording whether the annotator labeled the image correctly. We first compute the running accuracy of the annotator with a sliding window of 20 trials, and then rescale this curve to $T=50$ steps by averaging over consecutive blocks of four trials. The resulting value $p_t$ is the annotator's accuracy at step $t$.

We do this for three reasons. First, per-trial correctness is binary and noisy, so block averaging yields a smoother estimate of $p_t$ that preserves the drift pattern while removing trial-level fluctuations. Second, the 50,000 training images form only 250 sequences at length 200 but 1,000 sequences at length 50, which gives the rejector more independent sequences to learn from. Third, backpropagation through time for the recurrent rejector becomes four times longer at $T=200$, which substantially increases training cost across our searches and seeds. The rescaled curve thus fixes the sequence length $T = 50$ and the expert's accuracy $p_t$ at every step, which we use next to construct the training and test data.

\subsubsection{Data Construction}
\label{app:cifar10-dataset}

To match the $T = 50$ steps of the rescaled curve, we cut the 50{,}000 training images into consecutive sequences of 50 images and split them at the sequence level into training and validation sets (90\%/10\%), so that each sequence belongs entirely to one set. The image at position $t$ of a sequence is paired with the simulated expert at step $t$. During training, the expert's answers are redrawn every time a sequence is seen, so over 200 epochs each training sequence is paired with many different realisations of the expert. An incorrect expert answers with a uniformly random class, which can coincide with the label. The official 10{,}000-image \textsc{CIFAR-10} test set is never used during training and is reserved for evaluation.

At test time, a single test sequence would give a noisy estimate: the system accuracy at a step depends on which images appear and on whether the expert happens to be correct. We therefore estimate it by Monte Carlo replay. Each replay sequence draws $T = 50$ images uniformly at random, with replacement, from the test set, and pairs the image at step $t$ with an expert who is correct with probability $p_t$. We use 20{,}000 replay sequences for each of the 5 seeds, i.e.\ 100{,}000 per annotator, and all variants are evaluated on the same replay sequences. The Monte Carlo standard error of the accuracy at a single step is then at most $\sqrt{0.25/N}$: about $0.35$ percentage points for one seed ($N = 20{,}000$) and $0.16$ points when pooled over seeds ($N = 100{,}000$). This is small compared with the differences between variants, so the differences in our results reflect the trained models rather than evaluation noise.

\subsubsection{Hyperparameter Tuning}

For Naive L2D we tune the learning rate over $\{0.05, 0.1, 0.2\}$, training 5 seeds for 200 epochs at each value and selecting by the mean system accuracy on the held-out validation sequences (Appendix~\ref{app:cifar10-dataset}). $\mathrm{lr}=0.1$ is best for every annotator, and we use the same value for EDA-L2D, Per-step param., which differs from Naive L2D only in its defer outputs.

For EDA-L2D, RNN param., we search the learning rate jointly with the recurrent architecture: the cell (GRU or LSTM), the hidden size over $\{256, 512, 1024\}$, the number of layers over $\{1, 2\}$ and the learning rate over $\{0.01, 0.05, 0.1\}$. Every configuration is trained with 3 seeds for 200 epochs and evaluated on the validation images with 20{,}000 Monte Carlo sequences under realistic routing, as at test time (Table~\ref{tab:cifar10-rnn-search}). The learning rate matters most, whereas the architectures differ by less than one point at $\mathrm{lr}=0.1$. For each annotator we select the configuration with the highest validation system accuracy averaged over the seeds: a GRU with 256 units and 2 layers for annotator 454, an LSTM with 1024 units and 1 layer for annotator 1004, and an LSTM with 1024 units and 2 layers for annotator 54, all with $\mathrm{lr}=0.1$. The selected configuration is then trained with 5 seeds for the reported results.

\begin{table}[t]
\centering
\caption{Search over the recurrent architecture and learning rate of EDA-L2D, RNN param.\ on \textsc{CIFAR-10}. Validation system accuracy (\%) estimated with 20{,}000 Monte Carlo sequences per run; mean $\pm$ standard error over 3 seeds. $d$: hidden size; $L$: number of layers. Bold: the configuration selected for each annotator.}
\label{tab:cifar10-rnn-search}
\resizebox{\textwidth}{!}{%
\begin{tabular}{lcc ccc ccc ccc}
\toprule
 & & & \multicolumn{3}{c}{Annotator 454} & \multicolumn{3}{c}{Annotator 1004} & \multicolumn{3}{c}{Annotator 54} \\
\cmidrule(lr){4-6}\cmidrule(lr){7-9}\cmidrule(lr){10-12}
Cell & $d$ & $L$ & lr 0.1 & 0.05 & 0.01 & lr 0.1 & 0.05 & 0.01 & lr 0.1 & 0.05 & 0.01 \\
\midrule
\multirow{6}{*}{GRU}
 & 256 & 1 & 93.94\se{0.11} & 93.47\se{0.06} & 89.67\se{0.17} & 93.69\se{0.12} & 93.50\se{0.14} & 90.02\se{0.53} & 94.87\se{0.13} & 94.75\se{0.08} & 92.53\se{0.23} \\
 & 256 & 2 & \textbf{94.19}\se{0.27} & 93.60\se{0.15} & 89.27\se{0.31} & 93.92\se{0.28} & 93.53\se{0.33} & 90.38\se{0.34} & 95.00\se{0.32} & 94.63\se{0.20} & 92.52\se{0.22} \\
 & 512 & 1 & 93.94\se{0.05} & 92.92\se{0.33} & 89.58\se{0.15} & 93.85\se{0.06} & 93.68\se{0.06} & 89.55\se{0.30} & 94.90\se{0.05} & 94.59\se{0.14} & 92.28\se{0.30} \\
 & 512 & 2 & 93.56\se{0.28} & 92.91\se{0.35} & 89.53\se{0.20} & 93.82\se{0.02} & 93.46\se{0.17} & 90.18\se{0.12} & 94.99\se{0.07} & 93.92\se{0.25} & 92.28\se{0.09} \\
 & 1024 & 1 & 93.91\se{0.21} & 92.61\se{0.12} & 88.71\se{0.42} & 93.77\se{0.05} & 93.81\se{0.11} & 90.92\se{0.11} & 94.89\se{0.05} & 94.63\se{0.07} & 91.70\se{0.19} \\
 & 1024 & 2 & 93.64\se{0.22} & 92.75\se{0.52} & 88.60\se{0.27} & 93.92\se{0.02} & 93.76\se{0.07} & 88.85\se{0.42} & 95.04\se{0.18} & 94.34\se{0.09} & 91.98\se{0.09} \\
\midrule
\multirow{6}{*}{LSTM}
 & 256 & 1 & 94.02\se{0.10} & 93.83\se{0.08} & 91.12\se{0.20} & 93.97\se{0.08} & 93.60\se{0.11} & 91.03\se{0.22} & 95.06\se{0.07} & 94.87\se{0.10} & 93.00\se{0.10} \\
 & 256 & 2 & 93.94\se{0.17} & 93.52\se{0.19} & 90.77\se{0.15} & 93.82\se{0.13} & 93.37\se{0.21} & 90.78\se{0.05} & 94.99\se{0.18} & 94.93\se{0.16} & 92.79\se{0.06} \\
 & 512 & 1 & 93.87\se{0.18} & 93.61\se{0.15} & 90.76\se{0.18} & 93.70\se{0.19} & 93.53\se{0.08} & 90.49\se{0.33} & 95.03\se{0.16} & 94.84\se{0.13} & 92.72\se{0.20} \\
 & 512 & 2 & 93.85\se{0.23} & 93.50\se{0.25} & 90.49\se{0.40} & 93.55\se{0.19} & 93.38\se{0.16} & 90.69\se{0.16} & 95.05\se{0.15} & 94.85\se{0.18} & 92.55\se{0.22} \\
 & 1024 & 1 & 94.14\se{0.20} & 93.78\se{0.22} & 90.65\se{0.12} & \textbf{94.03}\se{0.08} & 93.66\se{0.11} & 90.76\se{0.08} & 95.15\se{0.05} & 95.13\se{0.07} & 92.77\se{0.08} \\
 & 1024 & 2 & 93.95\se{0.30} & 93.88\se{0.33} & 88.29\se{1.67} & 93.92\se{0.17} & 93.82\se{0.23} & 90.75\se{0.17} & \textbf{95.23}\se{0.16} & 95.10\se{0.18} & 92.14\se{0.26} \\
\bottomrule
\end{tabular}}
\end{table}

\subsection{CheXpert Setup}
\label{app:imple-chexpert}

\subsubsection{Data Construction}
\label{app:chexpert-dataset}
We use CheXpert-v1.0-small, the official downsampled-resolution release of CheXpert \citep{irvin2019chexpertlargechestradiograph}, and split its patients into 81\% training, 9\% validation and 10\% test. Unmentioned findings count as negative, and uncertain labels are excluded from the loss and the metrics. Within each set, the images are shuffled and cut into sequences, and the image at position $t$ is paired with the expert at step $t$. The expert follows the curve $a_t$ of Appendix~\ref{app:chexpert-drift} over a five-day week, with 5 steps per day for the general cohort and 10 for the emergency-department cohort. At step $t$, it answers each of the 14 findings correctly with probability $a_t$ and flips the label otherwise. Its answers are drawn once, so all variants see the same sequences and expert answers.

We consider two settings. In the \emph{whole-week} setting, one model is trained on week-long sequences of 25 or 50 steps. In the \emph{per-day} setting, a separate model is trained for each weekday on day-long sequences of 5 or 10 steps, with the expert following that day's part of the curve; each of these models uses all training images, and at test time their results are averaged over the five weekdays. In both settings, all variants are fine-tuned from the same classifier, pretrained without deferral on whole-week sequences.

\subsubsection{Hyperparameter Tuning}
\label{app:chexpert-hparams}
All variants use a DenseNet-121 initialised with ImageNet weights, with one class output and one defer output per finding, and are trained with Adam, learning rate $10^{-4}$ and weight decay $10^{-5}$. The classifier is first pretrained without deferral for 3 epochs, and every variant is then fine-tuned from it for 5 epochs; we use the final model. A batch contains 4 week-long or 20 day-long sequences, so every variant sees the same number of images per batch. 

EDA-L2D, Per-step param.\ has one defer head per step: 25 or 50 in the whole-week setting, 5 or 10 in the per-day setting. For EDA-L2D, RNN param., we search the recurrent cell over GRU and LSTM, the hidden size over $\{128, 256, 512\}$ and the number of layers over $\{1, 2, 3\}$. Each configuration is trained with two seeds on the general cohort and scored by the validation system AUROC after the last fine-tuning epoch. We select a GRU with hidden size 128 and 3 layers and use it for both cohorts and both settings. Its feedback input contains one correctness indicator per finding, set to $0.5$ before any outcome is observed.


\subsection{Hate Speech Setup}
\label{app:imple-hatespeech}

\subsubsection{Data Construction}
\label{app:hatespeech-dataset}
We use the dataset of \citet{davidson2017automatedhatespeechdetection}, with 24{,}783 tweets labelled as hate speech, offensive or neither. The tweets, in file order, form sequences of $T = 10$, and the complete sequences are split into 60\% training, 10\% validation and 30\% test. The tweet at position $t$ of a sequence is paired with the expert at step $t$, who is correct with probability $p_t$ given by one of the three synthetic curves of Appendix~\ref{app:hs-curves}: linear decline, warm-up then fatigue, and power-law practice. The expert's answers are drawn once when the data are built, so all variants are trained and evaluated on the same sequences and the same expert answers. EDA-L2D, Per-step param.\ trains one model per step; model $t$ is trained on all training tweets, with the expert's answers drawn from $p_t$.

\subsubsection{Hyperparameter Tuning}
\label{app:hatespeech-hparams}
All variants use a convolutional text encoder on 100-dimensional GloVe embeddings, with 300 filters of widths 3, 4 and 5 and dropout $0.5$; the vocabulary consists of the 25{,}000 most frequent tokens of the training tweets. We train with Adam and batch size 64, and keep the epoch with the highest validation system accuracy. Naive L2D and EDA-L2D, Per-step param.\ are trained for 10 epochs on batches of tweets. We select their learning rate from $\{2 \times 10^{-4}, 5 \times 10^{-4}, 10^{-3}, 2 \times 10^{-3}\}$ and use $10^{-3}$. EDA-L2D, RNN param.\ is trained for 20 epochs on batches of sequences. For each expert, we search the recurrent cell over GRU and LSTM, the learning rate over $\{5 \times 10^{-4}, 10^{-3}, 2 \times 10^{-3}\}$, the hidden size over $\{128, 256, 512\}$ and the number of layers over $\{1, 2\}$. An LSTM with learning rate $10^{-3}$, hidden size 512 and 2 layers performs well for all three experts, and we use it throughout. Its feedback input is the expert's correctness at the previous step, set to $0$ at the first step.

\section{Additional Results}
\subsection{CIFAR-10}
\label{app:cifar10-real-replay}

\begin{table}[htbp]
\centering
\caption{System accuracy (\%) on \textsc{CIFAR-10} when replaying the real answers of the drifting \textsc{CIFAR-10H} annotators: each annotator's 200 trials are split into 4 interleaved sequences of 50 steps, so that every trial keeps its original time step, and the accuracy is averaged over 5 seeds. In brackets: the gain over Naive L2D with the 95\% interval of a paired block bootstrap over trials (blocks of 10 trials, $10{,}000$ resamples); all methods are evaluated on the same trials.}
\label{tab:cifar10-real-answer}
\begin{threeparttable}
\small
\begin{tabular*}{\textwidth}{@{\extracolsep{\fill}}lccc}
\toprule
Method & Annotator 454 & Annotator 1004 & Annotator 54 \\
\midrule
Naive L2D & $87.7$ & $89.0$ & $95.3$ \\
EDA-L2D, Per-step param. & $92.2$ {\scriptsize$(\mathbf{+4.5}\,[2.6, 6.4])$} & $90.7$ {\scriptsize$(+1.7\,[0.0, 3.4])$} & $95.6$ {\scriptsize$(+0.3\,[-1.1, 1.7])$} \\
EDA-L2D, RNN param. & $91.5$ {\scriptsize$(\mathbf{+3.8}\,[1.7, 6.0])$} & $90.8$ {\scriptsize$(\mathbf{+1.8}\,[0.5, 3.3])$} & $96.1$ {\scriptsize$(+0.8\,[-0.3, 1.9])$} \\
\quad delay 10\tnote{a} & $91.1$ {\scriptsize$(\mathbf{+3.4}\,[1.0, 5.7])$} & $91.0$ {\scriptsize$(\mathbf{+2.0}\,[0.7, 3.5])$} & $96.2$ {\scriptsize$(+0.9\,[-0.1, 1.9])$} \\
\quad no feedback\tnote{b} & $91.2$ {\scriptsize$(\mathbf{+3.5}\,[1.3, 5.7])$} & $91.0$ {\scriptsize$(\mathbf{+2.0}\,[0.7, 3.5])$} & $96.0$ {\scriptsize$(+0.7\,[-0.3, 1.7])$} \\
\bottomrule
\end{tabular*}
\begin{tablenotes}\footnotesize
\item[a] The observed outcomes reach the RNN only every 10 steps.
\item[b] The RNN receives no feedback about the expert's correctness.
\end{tablenotes}
\end{threeparttable}
\end{table}

\subsection{CheXpert}
\label{app:chexpert-additional-results}

\begin{figure}[htbp]
\centering
\includegraphics[width=0.6\textwidth, trim=0 6 0 0, clip]{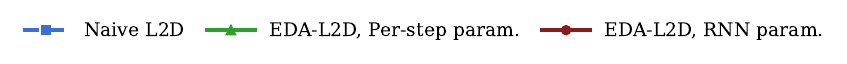}\\[-2pt]
\begin{minipage}[t]{0.495\textwidth}
  \centering
  {\scriptsize General cohort}\\[1pt]
  \includegraphics[width=\linewidth]{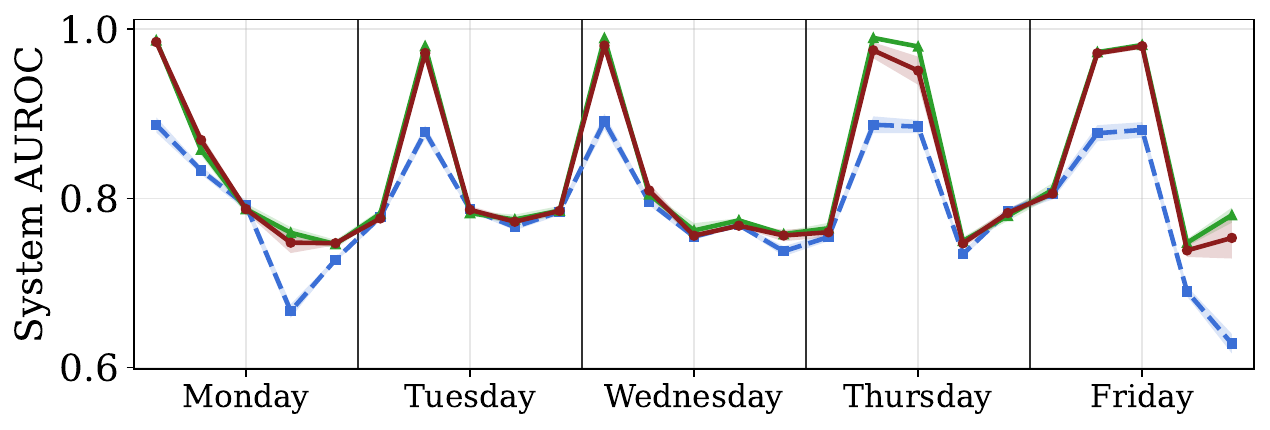}\\[2pt]
  \includegraphics[width=\linewidth]{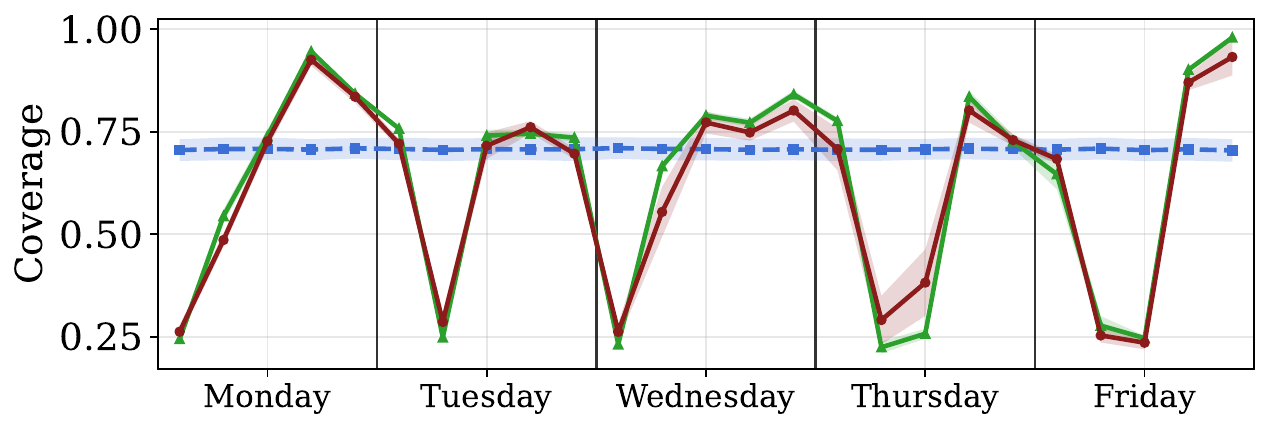}
\end{minipage}\hfill
\begin{minipage}[t]{0.495\textwidth}
  \centering
  {\scriptsize Emergency-department cohort}\\[1pt]
  \includegraphics[width=\linewidth]{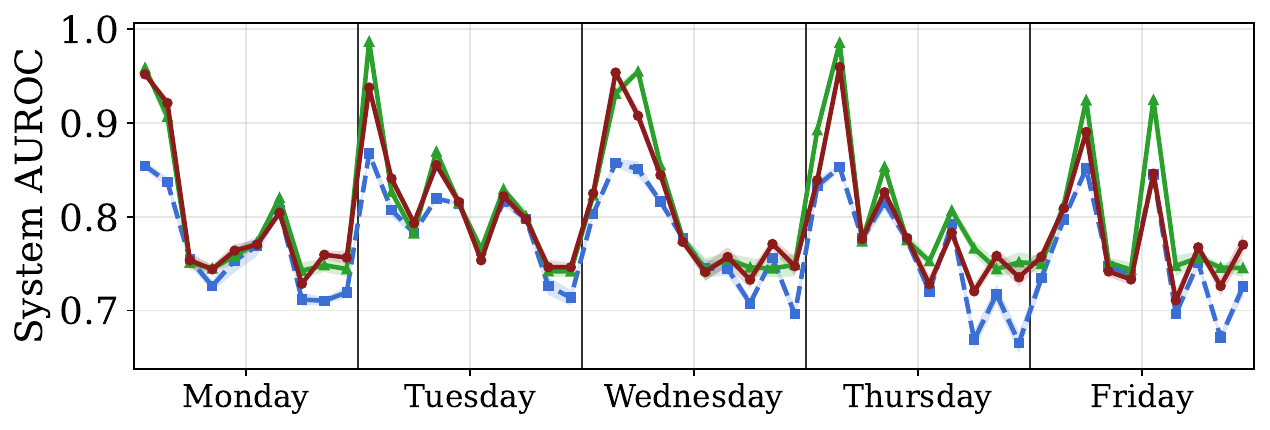}\\[2pt]
  \includegraphics[width=\linewidth]{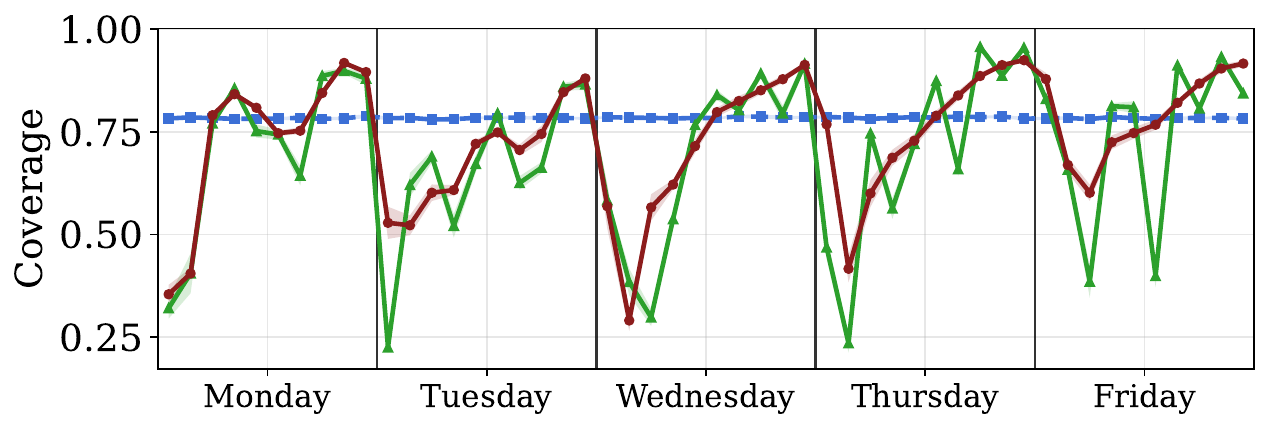}
\end{minipage}
\caption{CheXpert with one model for the whole week, for the general cohort (left) and the emergency-department cohort (right). Top: system AUROC, averaged over the 14 findings; bottom: coverage. Vertical lines separate the weekdays. Lines show the mean over 5 seeds and shaded bands $\pm$ one standard error.}
\label{fig:chexpert-week}
\end{figure}

\paragraph{One model for the whole week.}
In the whole-week setting, a single model is trained on week-long sequences and must cover all five days (Tables~\ref{tab:chexpert-week-general} and~\ref{tab:chexpert-week-emergency}, Figure~\ref{fig:chexpert-week}). Both parameterizations of EDA-L2D again outperform Naive L2D in both cohorts, raising the average system AUROC from $0.791$ to $0.835$ and $0.830$ in the general cohort and from $0.770$ to $0.804$ and $0.795$ in the emergency-department cohort. The gains are close to those of the per-day setting for EDA-L2D, Per-step param., while EDA-L2D, RNN param.\ gains somewhat less, most visibly in the emergency-department cohort, where it must follow a longer sequence of 50 steps.

\begin{table}[htbp]
\centering
\caption{System AUROC on CheXpert, general cohort, one model for the whole week, for three findings and averaged over all 14 findings; each value is averaged over the steps of the week, with mean $\pm$ standard error over 5 seeds. EDA-L2D, RNN param.\ is evaluated under realistic routing.}
\label{tab:chexpert-week-general}
\small
\begin{tabular*}{\textwidth}{@{\extracolsep{\fill}}lcccc}
\toprule
Method & Enl.\ Cardiomediastinum & Consolidation & Fracture & Average (14 findings) \\
\midrule
Naive L2D & $0.681 \pm 0.007$ & $0.781 \pm 0.010$ & $0.716 \pm 0.006$ & $0.791 \pm 0.002$ \\
EDA-L2D, Per-step param. & $\mathbf{0.755 \pm 0.004}$ & $\mathbf{0.822 \pm 0.005}$ & $\mathbf{0.792 \pm 0.009}$ & $\mathbf{0.835 \pm 0.001}$ \\
EDA-L2D, RNN param. & $\mathbf{0.748 \pm 0.010}$ & $\mathbf{0.823 \pm 0.006}$ & $\mathbf{0.789 \pm 0.012}$ & $0.830 \pm 0.003$ \\
\bottomrule
\end{tabular*}
\end{table}

\begin{table}[htbp]
\centering
\caption{System AUROC on CheXpert, emergency-department cohort, one model for the whole week; same layout as Table~\ref{tab:chexpert-week-general}.}
\label{tab:chexpert-week-emergency}
\small
\begin{tabular*}{\textwidth}{@{\extracolsep{\fill}}lcccc}
\toprule
Method & Enl.\ Cardiomediastinum & Consolidation & Fracture & Average (14 findings) \\
\midrule
Naive L2D & $0.664 \pm 0.003$ & $0.738 \pm 0.003$ & $0.716 \pm 0.008$ & $0.770 \pm 0.002$ \\
EDA-L2D, Per-step param. & $\mathbf{0.720 \pm 0.011}$ & $\mathbf{0.791 \pm 0.004}$ & $\mathbf{0.765 \pm 0.009}$ & $\mathbf{0.804 \pm 0.002}$ \\
EDA-L2D, RNN param. & $0.701 \pm 0.003$ & $0.783 \pm 0.003$ & $0.739 \pm 0.009$ & $0.795 \pm 0.002$ \\
\bottomrule
\end{tabular*}
\end{table}

\begin{figure}[htbp]
\centering
\includegraphics[width=0.6\textwidth, trim=0 6 0 0, clip]{figures/chexpert/general_dayamp_day_legend.pdf}\\[-2pt]
\begin{minipage}[t]{0.495\textwidth}
  \centering
  {\scriptsize General cohort}\\[1pt]
  \includegraphics[width=\linewidth]{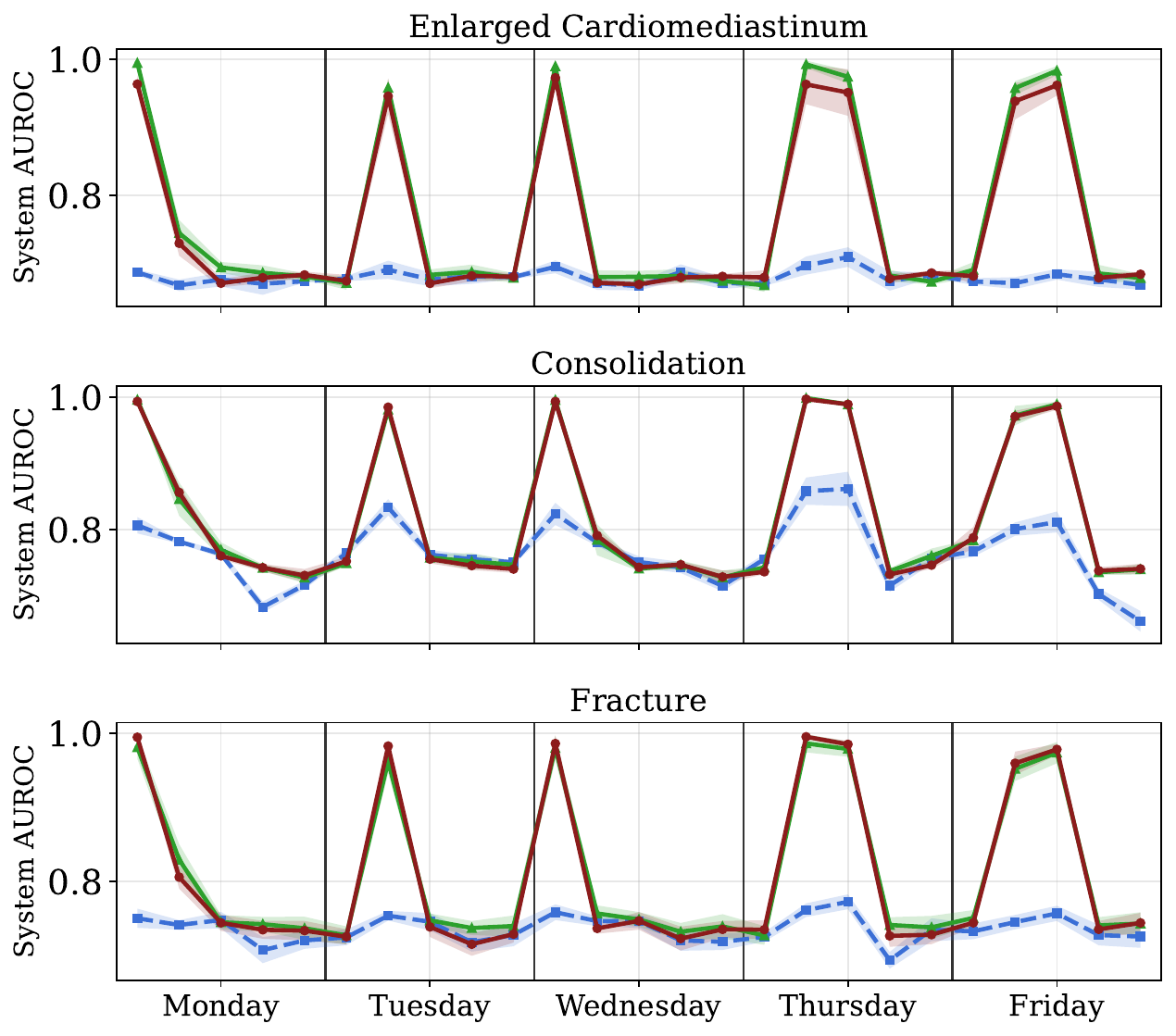}
\end{minipage}\hfill
\begin{minipage}[t]{0.495\textwidth}
  \centering
  {\scriptsize Emergency-department cohort}\\[1pt]
  \includegraphics[width=\linewidth]{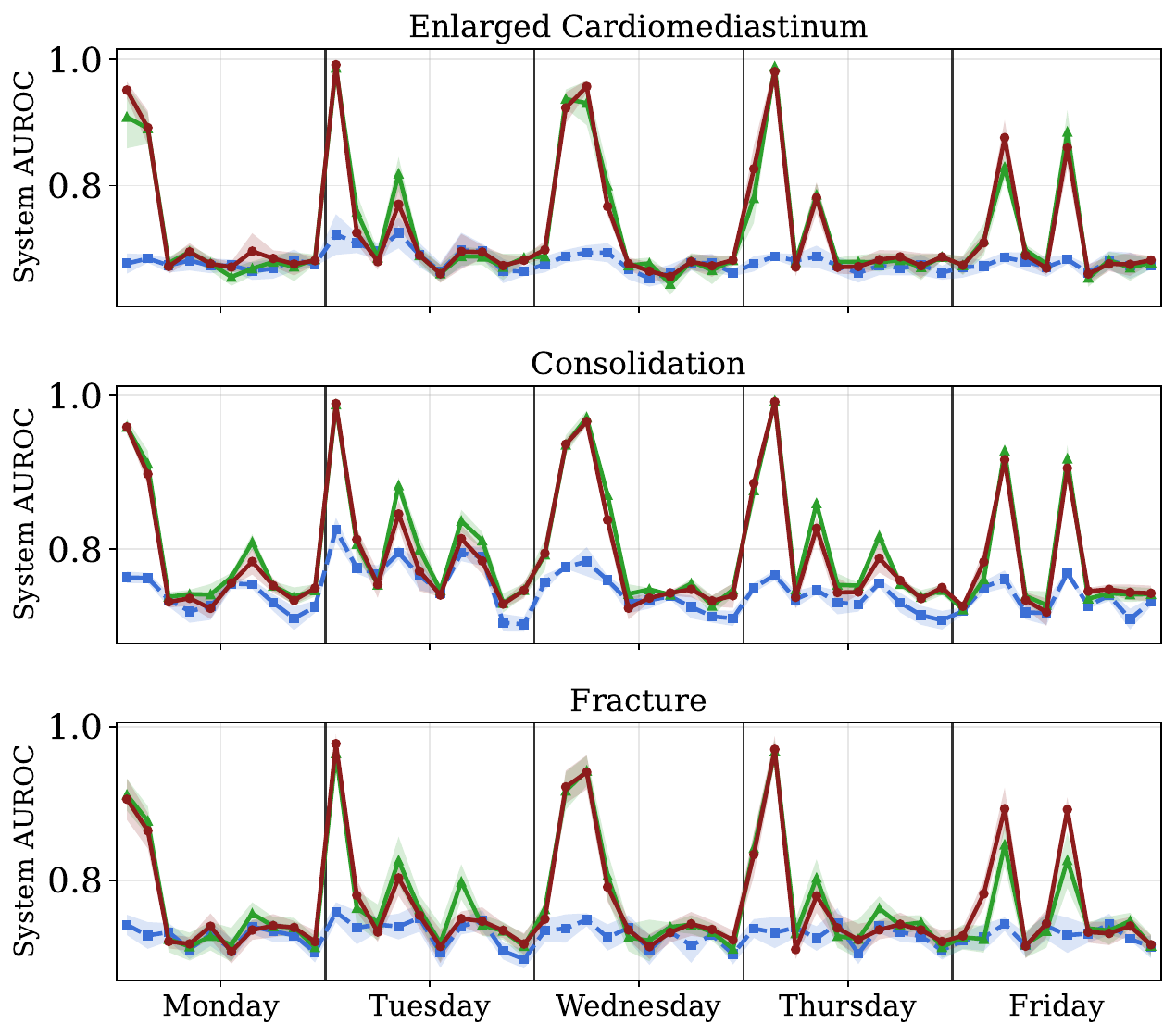}
\end{minipage}
\caption{System AUROC along the week for three findings, CheXpert with one model per weekday, for the general cohort (left) and the emergency-department cohort (right). Lines show the mean over 5 seeds and shaded bands $\pm$ one standard error.}
\label{fig:chexpert-findings-day}
\end{figure}

\begin{figure}[htbp]
\centering
\includegraphics[width=0.6\textwidth, trim=0 6 0 0, clip]{figures/chexpert/general_dayamp_week_legend.pdf}\\[-2pt]
\begin{minipage}[t]{0.495\textwidth}
  \centering
  {\scriptsize General cohort}\\[1pt]
  \includegraphics[width=\linewidth]{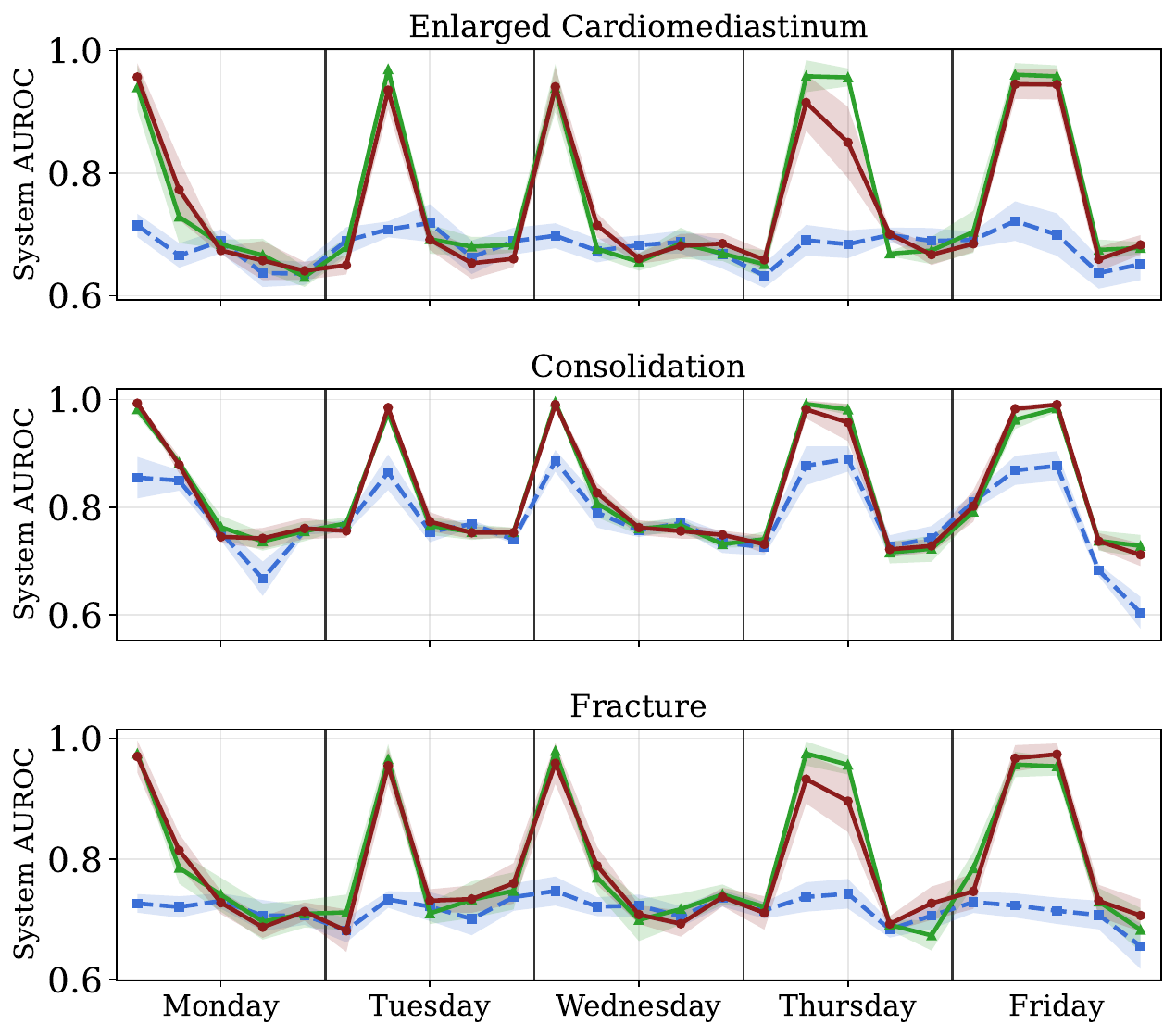}
\end{minipage}\hfill
\begin{minipage}[t]{0.495\textwidth}
  \centering
  {\scriptsize Emergency-department cohort}\\[1pt]
  \includegraphics[width=\linewidth]{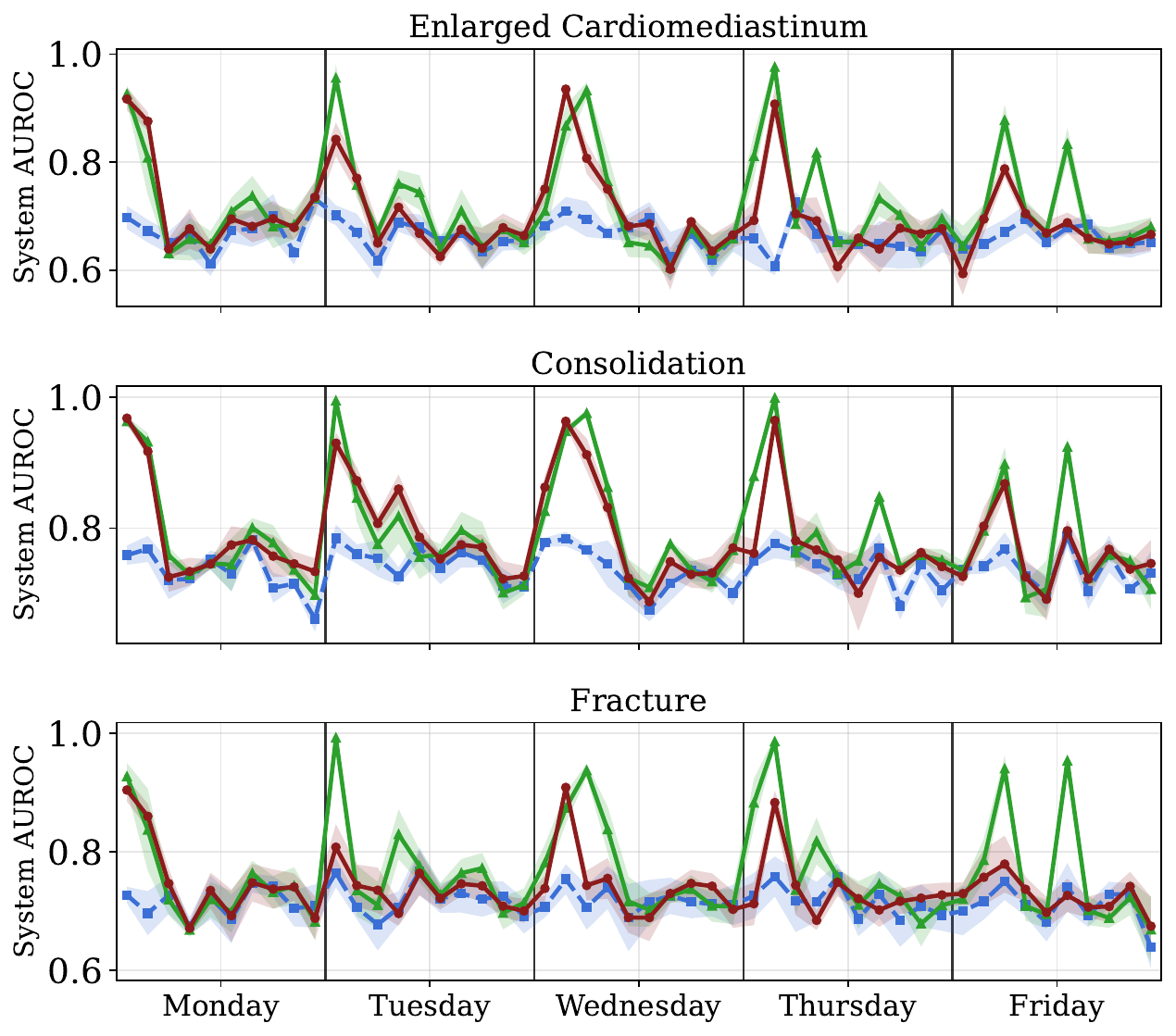}
\end{minipage}
\caption{System AUROC along the week for three findings, CheXpert with one model for the whole week, for the general cohort (left) and the emergency-department cohort (right). Lines show the mean over 5 seeds and shaded bands $\pm$ one standard error.}
\label{fig:chexpert-findings-week}
\end{figure}

\paragraph{Individual findings.}
Figures~\ref{fig:chexpert-findings-day} and~\ref{fig:chexpert-findings-week} show the system AUROC along the week for three findings on which the classifier alone is weak: Enlarged Cardiomediastinum, Consolidation and Fracture. In both settings and both cohorts, EDA-L2D reaches a system AUROC close to $1$ at the steps where the expert is most accurate, whereas Naive L2D stays close to the classifier alone, which accounts for the large per-finding gains in Tables~\ref{tab:chexpert-day-general}--\ref{tab:chexpert-week-emergency}.

\section{Robustness Evaluation}
\label{app:robustness}
In the main experiments, the expert at test time follows the same drift as during training. In practice, however, the expert's drift at deployment may differ from the drift observed during model training. We therefore evaluate how the trained models behave when the expert departs from the training drift. All models are trained on the original drift, as in the main experiments, and are \emph{not} retrained. At test time, we change only the expert or the feedback indicating whether the expert's decisions were correct. Wherever possible, we use the same random draws for the expert so that the perturbed and unperturbed runs, as well as all methods within each run, are paired. In every test, we compare Naive L2D with EDA-L2D using either the Per-step or RNN parameterization.

\subsection{Imperfect feedback}
\label{app:robust-imperfect-fb}
EDA-L2D, RNN param.\ is trained with the expert's correctness at every step, observed immediately and without error. At deployment, this feedback is far more limited. Under realistic routing, the model observes the expert's correctness only for deferred inputs and otherwise retains the most recent observed outcome. Keeping the expert's original accuracy curve, we degrade this feedback further in three ways:
\begin{itemize}
    \item \emph{Delay}: observed outcomes reach the model only once every $k\in\{5,10\}$ steps, so the feedback can be up to $k$ steps old.
    \item \emph{Noise}: each observed outcome is flipped with probability $\varepsilon\in\{0.1,0.2,0.3\}$.
    \item \emph{Partial observation}: a deferred input reveals its outcome only with probability $\pi\in\{0.5,0.25\}$.
\end{itemize}
We measure the contribution of feedback as the gain in system accuracy over the same model evaluated without feedback.

Table~\ref{tab:cifar10-robust-feedback} shows that this contribution degrades gracefully, and its mean remains positive under every condition and for every annotator. Delay is the most harmful degradation. For annotator 454, delivering outcomes only every 10 steps reduces the contribution from $+0.36$ to $+0.15$ points, whereas flipping 30\% of the outcomes or revealing only a quarter of them still retains roughly 70--75\% of it ($+0.27$ and $+0.25$ points). This ordering is expected: on a drifting expert, a stale outcome misrepresents the expert's current state, while an occasional noisy or missing outcome is averaged out by the recurrent state. For annotators 1004 and 54, the contribution stays below $0.05$ points in every condition.

These contributions are an order of magnitude smaller than the overall gain of EDA-L2D, RNN param.\ over Naive L2D ($+3.32$, $+2.15$ and $+0.92$ points; Table~\ref{tab:cifar10}). Most of the advantage on a fixed drift curve therefore comes from the temporal pattern learned during training. This is unsurprising, because when the expert follows the same curve at training and test time, the time step alone is highly informative about their accuracy, leaving little for feedback to add. Feedback becomes more valuable when the expert departs from this pattern: during a temporary accuracy dip, its contribution grows with the depth of the dip (Appendix~\ref{app:robust-dip}). Taken together, feedback provides a small but consistent additional gain on the original curve, and this gain does not depend on observations being immediate, accurate or complete.

\begin{table}[htbp]
\centering
\caption{Imperfect feedback on \textsc{CIFAR-10} (original expert curve). Gain in system accuracy (percentage points) of the RNN under each feedback condition over the same RNN without feedback, over all 50 steps; mean $\pm$ s.e.\ over 5 seeds.}
\label{tab:cifar10-robust-feedback}
\small
\begin{tabular}{lccc}
\toprule
Feedback condition & Annotator 454 & Annotator 1004 & Annotator 54 \\
\midrule
Realistic routing (main setting) & $+0.360 \pm 0.213$ & $+0.032 \pm 0.002$ & $+0.047 \pm 0.014$ \\
\midrule
Delay $k = 5$                    & $+0.296 \pm 0.142$ & $+0.022 \pm 0.002$ & $+0.047 \pm 0.013$ \\
Delay $k = 10$                   & $+0.150 \pm 0.076$ & $+0.020 \pm 0.002$ & $+0.042 \pm 0.010$ \\
\midrule
Noise $\varepsilon = 0.1$        & $+0.333 \pm 0.193$ & $+0.028 \pm 0.002$ & $+0.045 \pm 0.013$ \\
Noise $\varepsilon = 0.2$        & $+0.303 \pm 0.172$ & $+0.028 \pm 0.001$ & $+0.041 \pm 0.011$ \\
Noise $\varepsilon = 0.3$        & $+0.272 \pm 0.150$ & $+0.024 \pm 0.002$ & $+0.038 \pm 0.010$ \\
\midrule
Partial $\pi = 0.5$              & $+0.324 \pm 0.177$ & $+0.030 \pm 0.002$ & $+0.045 \pm 0.013$ \\
Partial $\pi = 0.25$             & $+0.252 \pm 0.118$ & $+0.033 \pm 0.003$ & $+0.039 \pm 0.009$ \\
\bottomrule
\end{tabular}
\end{table}

\subsection{Temporary dip}
\label{app:robust-dip}
The expert's accuracy drops by $d \in \{0.05, 0.10, 0.15\}$ for 5 consecutive steps and then returns to its curve. The dip starts at step $1, 6, \ldots, 46$, and we average over these 10 positions. We report system accuracy during the dip and in the 5 steps after it.

\begin{figure}[htbp]
\centering
\includegraphics[width=\textwidth]{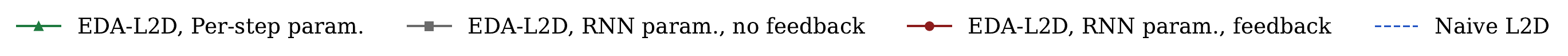}\\[2pt]
\includegraphics[width=\textwidth]{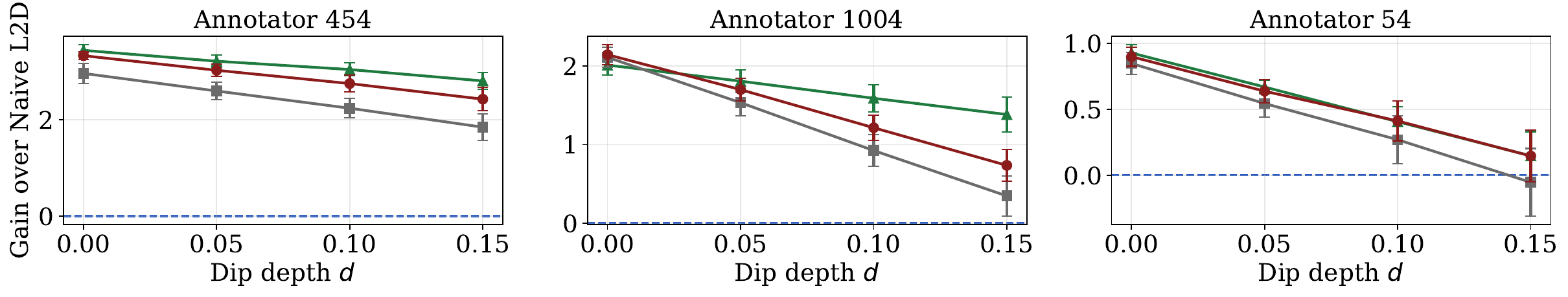}
\caption{Temporary dip on \textsc{CIFAR-10}: gain in system accuracy over Naive L2D (percentage points) during the dip, as a function of the dip depth $d$, averaged over the 10 dip positions. Error bars show $\pm$ one standard error over 5 seeds; the dashed line marks Naive L2D.}
\label{fig:cifar10-robust-dip}
\end{figure}

During the dip, both EDA-L2D variants stay above Naive L2D at every dip depth for all three annotators (Table~\ref{tab:cifar10-robust-dip}). Their gain shrinks as the dip deepens, because a model that relies on the expert's usual accuracy defers too often while the expert is worse (Figure~\ref{fig:cifar10-robust-dip}). For EDA-L2D, RNN param., feedback reduces this loss, and its contribution grows with the depth of the dip: from $+0.37$, $+0.03$ and $+0.05$ points without a dip to $+0.58$, $+0.39$ and $+0.20$ points at $d = 0.15$ for annotators 454, 1004 and 54 (Table~\ref{tab:cifar10-robust-dip-fb}). Without feedback, the model falls slightly below Naive L2D for annotator 54 at $d = 0.15$, whereas with feedback it stays above. In the five steps after the dip, the gains of both variants return to their level without a dip, so a temporary drop does not affect later decisions (Table~\ref{tab:cifar10-robust-dip-after}). These results also show that feedback helps EDA-L2D, RNN param.\ when the expert's accuracy temporarily drops below the level seen in training.

\begin{table}[htbp]
\centering
\caption{Temporary dip of depth $d$ for 5 steps on \textsc{CIFAR-10}: gain in system accuracy over Naive L2D (percentage points) during the dip, averaged over the 10 dip positions ($d = 0$: no dip, same steps); mean $\pm$ s.e.\ over 5 seeds.}
\label{tab:cifar10-robust-dip}
\small
\begin{tabular*}{\textwidth}{@{\extracolsep{\fill}}llcccc}
\toprule
Annotator & Method & $d = 0$ & $d = 0.05$ & $d = 0.10$ & $d = 0.15$ \\
\midrule
\multirow{2}{*}{454}  & EDA-L2D, Per-step param. & $+3.44$\se{0.11} & $+3.21$\se{0.13} & $+3.04$\se{0.13} & $+2.80$\se{0.18} \\
                      & EDA-L2D, RNN param.      & $+3.33$\se{0.07} & $+3.02$\se{0.13} & $+2.75$\se{0.17} & $+2.43$\se{0.24} \\
\midrule
\multirow{2}{*}{1004} & EDA-L2D, Per-step param. & $+2.01$\se{0.13} & $+1.81$\se{0.15} & $+1.59$\se{0.18} & $+1.38$\se{0.22} \\
                      & EDA-L2D, RNN param.      & $+2.15$\se{0.13} & $+1.70$\se{0.14} & $+1.22$\se{0.16} & $+0.74$\se{0.20} \\
\midrule
\multirow{2}{*}{54}   & EDA-L2D, Per-step param. & $+0.93$\se{0.06} & $+0.67$\se{0.05} & $+0.41$\se{0.11} & $+0.15$\se{0.18} \\
                      & EDA-L2D, RNN param.      & $+0.90$\se{0.07} & $+0.64$\se{0.09} & $+0.41$\se{0.15} & $+0.15$\se{0.20} \\
\bottomrule
\end{tabular*}
\end{table}

\begin{table}[htbp]
\centering
\caption{Temporary dip on \textsc{CIFAR-10}: gain in system accuracy over Naive L2D (percentage points) in the 5 steps after the dip, averaged over the dip positions; mean $\pm$ s.e.\ over 5 seeds.}
\label{tab:cifar10-robust-dip-after}
\small
\begin{tabular*}{\textwidth}{@{\extracolsep{\fill}}llccc}
\toprule
Annotator & Method & $d = 0.05$ & $d = 0.10$ & $d = 0.15$ \\
\midrule
\multirow{2}{*}{454}  & EDA-L2D, Per-step param. & $+3.54$\se{0.11} & $+3.57$\se{0.12} & $+3.58$\se{0.13} \\
                      & EDA-L2D, RNN param.      & $+3.38$\se{0.07} & $+3.40$\se{0.08} & $+3.39$\se{0.07} \\
\midrule
\multirow{2}{*}{1004} & EDA-L2D, Per-step param. & $+2.05$\se{0.13} & $+2.07$\se{0.14} & $+2.12$\se{0.13} \\
                      & EDA-L2D, RNN param.      & $+2.21$\se{0.14} & $+2.23$\se{0.13} & $+2.27$\se{0.13} \\
\midrule
\multirow{2}{*}{54}   & EDA-L2D, Per-step param. & $+0.95$\se{0.06} & $+0.93$\se{0.06} & $+0.93$\se{0.07} \\
                      & EDA-L2D, RNN param.      & $+0.94$\se{0.08} & $+0.90$\se{0.08} & $+0.90$\se{0.08} \\
\bottomrule
\end{tabular*}
\end{table}

\begin{table}[htbp]
\centering
\caption{Temporary dip on \textsc{CIFAR-10}: gain in system accuracy (percentage points) of EDA-L2D, RNN param.\ over the same model without feedback, during the dip; mean $\pm$ s.e.\ over 5 seeds.}
\label{tab:cifar10-robust-dip-fb}
\small
\begin{tabular*}{\textwidth}{@{\extracolsep{\fill}}lcccc}
\toprule
Annotator & $d = 0$ & $d = 0.05$ & $d = 0.10$ & $d = 0.15$ \\
\midrule
454  & $+0.37$\se{0.21} & $+0.43$\se{0.15} & $+0.51$\se{0.11} & $+0.58$\se{0.10} \\
1004 & $+0.03$\se{0.00} & $+0.17$\se{0.02} & $+0.29$\se{0.04} & $+0.39$\se{0.06} \\
54   & $+0.05$\se{0.01} & $+0.09$\se{0.03} & $+0.14$\se{0.06} & $+0.20$\se{0.10} \\
\bottomrule
\end{tabular*}
\end{table}

\subsection{Beyond the training horizon}
\label{app:robust-horizon}
All models are trained on sequences of 50 steps. To test whether they still work after the last trained step, we extend the sequences to 60 steps: steps 1--50 follow the original curve, and in steps 51--60 the expert keeps its accuracy at step 50, shifted by $\Delta \in \{0, \pm 0.05, \pm 0.10\}$. EDA-L2D, Per-step param.\ has no defer head beyond step 50 and reuses its last one, while EDA-L2D, RNN param.\ continues its recurrence. We report system accuracy over steps 51--60.

For annotator 454, EDA-L2D, RNN param.\ remains well above Naive L2D beyond the training horizon, by $+5.71$ to $+8.17$ points for every $\Delta$ (Table~\ref{tab:cifar10-robust-horizon}), so the benefit of modelling the expert's drift carries over to steps it was not trained on. For annotators 1004 and 54, the gain is small and depends on the direction of the change: the model stays above Naive L2D when the expert's accuracy decreases for annotator 1004 and when it increases for annotator 54, but falls below it when the change goes the other way. EDA-L2D, Per-step param., which reuses its last head, stays above Naive L2D in every setting and is the more robust choice beyond the training horizon.

\begin{table}[htbp]
\centering
\caption{Beyond the training horizon on \textsc{CIFAR-10}: in steps 51--60 the expert keeps its accuracy at step 50, shifted by $\Delta$. Gain in system accuracy over Naive L2D (percentage points) over steps 51--60, mean $\pm$ s.e.\ over 5 seeds.}
\label{tab:cifar10-robust-horizon}
\small
\begin{tabular*}{\textwidth}{@{\extracolsep{\fill}}llccccc}
\toprule
Annotator & Method & $\Delta = -0.10$ & $-0.05$ & $0$ & $+0.05$ & $+0.10$ \\
\midrule
\multirow{2}{*}{454}  & EDA-L2D, Per-step param. & $+9.35$\se{0.35} & $+8.51$\se{0.32} & $+7.84$\se{0.36} & $+7.17$\se{0.31} & $+6.46$\se{0.29} \\
                      & EDA-L2D, RNN param.      & $+8.17$\se{0.22} & $+7.52$\se{0.19} & $+6.90$\se{0.18} & $+6.31$\se{0.18} & $+5.71$\se{0.16} \\
\midrule
\multirow{2}{*}{1004} & EDA-L2D, Per-step param. & $+0.81$\se{0.14} & $+0.81$\se{0.16} & $+0.86$\se{0.13} & $+0.84$\se{0.14} & $+0.83$\se{0.14} \\
                      & EDA-L2D, RNN param.      & $+0.52$\se{0.12} & $+0.25$\se{0.14} & $+0.00$\se{0.13} & $-0.28$\se{0.12} & $-0.52$\se{0.14} \\
\midrule
\multirow{2}{*}{54}   & EDA-L2D, Per-step param. & $+0.20$\se{0.04} & $+0.24$\se{0.09} & $+0.31$\se{0.16} & $+0.26$\se{0.24} & $+0.29$\se{0.25} \\
                      & EDA-L2D, RNN param.      & $-1.58$\se{0.61} & $-0.73$\se{0.37} & $+0.10$\se{0.22} & $+0.90$\se{0.19} & $+0.96$\se{0.23} \\
\bottomrule
\end{tabular*}
\end{table}

\end{document}